\documentclass[journal]{IEEEtran}
\usepackage{amsmath,amsfonts}
\usepackage{algorithmic}
\usepackage{algorithm}
\usepackage{array}
\usepackage{textcomp}
\usepackage{stfloats}
\usepackage{url}
\usepackage{verbatim}
\usepackage{graphicx}
\usepackage{cite}
\usepackage{bbm}
\usepackage[justification=centering]{caption}
\usepackage{booktabs}
\usepackage{multirow}
\usepackage{makecell}
\usepackage{subfigure}
\usepackage[numbers,sort&compress]{natbib}
\def\BibTeX{{\rm B\kern-.05em{\sc i\kern-.025em b}\kern-.08em
    T\kern-.1667em\lower.7ex\hbox{E}\kern-.125emX}}
\usepackage{balance}
\usepackage{multicol}

\begin{document}

\title{Pursuing 3D Scene Structures with Optical Satellite Images from Affine Reconstruction to Euclidean Reconstruction}

\author{Pinhe~Wang,
		Limin~Shi,
        Bao~Chen,
        Zhanyi~Hu,
        Qiulei~Dong,
        and~Jianzhong~Qiao
\thanks{Pinhe Wang and Jianzhong Qiao are with the School of Computer Science and Engineering, University of Northeastern, Shenyang 110169, China.} 
\thanks{Limin Shi is with the Research Center of Aerospace Information, Institute of Automation, Chinese Academy of Sciences, Beijing, 100190, China.} 
\thanks{Bao Chen, Zhanyi Hu and Qiulei Dong are with the National Laboratory of Pattern Recongnition, Institute of Automation, Chinese Academy of Sciences, Beijing, 100190, China, also with the School of Artificial Intelligence, University of Chinese Academy of Sciences, Beijing 100049, China, and also with the Center for Excellence in Brain Science and Intelligence Technologym Chinese Academy of Sciences, Beijing 100190, China (e-mail: qldong@nlpr.ia.ac.cn)(Corresponding author: Qiulei Dong).} 
}

\maketitle

\begin{abstract}
How to use multiple optical satellite images to recover the 3D scene structure is a challenging and important problem in the remote sensing field. Most existing methods in literature have been explored based on the classical RPC (rational polynomial camera) model which requires at least 39 GCPs (ground control points), however, it is not trivial to obtain such a large number of GCPs in many real scenes. Addressing this problem, we propose a hierarchical reconstruction framework based on multiple optical satellite images, which needs only 4 GCPs. The proposed framework is composed of an affine dense reconstruction stage and a followed affine-to-Euclidean upgrading stage: At the affine dense reconstruction stage, an affine dense reconstruction approach is explored for pursuing the 3D affine scene structure without any GCP from input satellite images. Then at the affine-to-Euclidean upgrading stage, the obtained 3D affine structure is upgraded to a Euclidean one with 4 GCPs. Experimental results on two public datasets demonstrate that the proposed method significantly outperforms three state-of-the-art methods in most cases.\cite{1985STEREO} 
\end{abstract}

\begin{IEEEkeywords}
Optical satellite image, Affine dense reconstrution, RPC, GCPs, Affine-to-Euclidean upgrading.
\end{IEEEkeywords}

\IEEEpeerreviewmaketitle

\section{Introduction}
\IEEEPARstart{W}{ith} the rapid development of the aerospace photography technology, an increasing number of high-resolution optical satellite images have been captured. In recent years, 3D scene reconstruction with optical satellite images, which aims to utilize multiple 2D scene images to automatically recover the corresponding 3D scene structure, has attracted more and more attention in both the fields of remote sensing and computer vision \cite{2019MULTI, 2020HP,2020QNORM,2021GCP,2021L1,2021NBC,2021SEMANTIC,2012COLLINEARITY,2012CARTOSAT,2012MULTIVIEW,2013LARGESCALE,2014S2P,2016BENCHMARK,2016ASP,2018ASP,2019COLMAP}.\\
\indent
It is worth noting that although a large number of 3D reconstruction methods have been proposed in the computer vision field \cite{2011CMPMVS,2014MVE,2016COLMAP,2018STRATIFIED,2019NSFM,2020TERE}, most of them are generally available to handle the images captured by a plane array camera, but fail to straightforwardly handle the satellite images captured by a pushbroom camera where a linear sensor array is utilized to capture a line of imagery at a time, due to the fact that the imaging model of a linear pushbroom camera is quite distinct from that of a plane array camera \cite{1987COLLINEARITY,1988COLLINEARITY, 2000LPCE,2005LPCE}. In the remote sensing field, existing sensor models are roughly divided in two categories \cite{2001RPC}: physical models and generalized models. A physical model only represents the physical imaging process of a special sensor, but it is not available for handling the data from other sensors \cite{2005COLLINEARITY}. Generalized models utilize some general functions (e.g. polynomial function, rational function, etc.) to represent the functional relationship between the image space and the scene space, and they are able to effectively handle various sensor data \cite{1997RPC, 1997LPC}. Hence, most of the recent 3D reconstruction methods with satellite optical images employ a generalized sensor model.\\
\indent
The RPC (rational polynomial camera) model \cite{1997RPC} is a popular generalized sensor model, which is extensively used for 3D reconstruction with optical satellite images in literature \cite{2012CARTOSAT, 2014S2P, 2016BENCHMARK}. Theoretically, it requires at least 39 uniformly distributed GCPs (ground control points) with accurate geographical coordinates to fit the RPC model, this is to say, these RPC-based reconstructions methods also have to use at least 39 GCPs during their reconstruction processes. However, since the scenes involved in optical satellite images are generally huge, it has to take a lot of time and effort to collect such GCPs, and the bigger the number of the collected GCPs is, the more laborious the GCPs collection is. This issue naturally raises the following problem: how to reconstruct 3D scenes based on optical satellite images by utilizing as few GCPs as possible?\\
\indent
Addressing this problem, we propose a hierarchical reconstruction framework based on multiple optical satellite images, which reconstructs the affine and Euclidean scene structures sequentially, called AE-Rec. At the first stage, we explore an affine dense reconstruction approach for obtaining the 3D affine structure from input satellite images, assuming that local small-sized tiles in satellite images is approximately subject to an affine camera model. This explored affine approach performs under a designed incremental reconstruction strategy, and it does not use any GCP. At the second stage, the obtained 3D affine structure is upgraded to a Euclidean one by fitting a global transformation matrix with at least 4 GCPs.\\
\indent
In sum, the main contributions in this paper include:
\begin{itemize}
	\item[1] We explore a dense 3D affine reconstruction approach based on multiple satellite images, which employs an incremental reconstruction strategy to reconstruct affine scene structures without any GCP.
	\item[2] We propose the AE-Rec framework for 3D Euclidean reconstruction based on optical satellite images, which could naturally accommodate the aforementioned 3D affine reconstruction approach. The proposed AE-Rec requires at least 4 GCPs, far less than the required minimum GCP number 39 by the existing RPC-based methods in literature \cite{2012CARTOSAT, 2014S2P, 2016BENCHMARK}. Experimental results in Section IV further demonstrate that even if only 4 GCPs are used in the proposed AE-Rec, it still performs better than three comparative RPC-based methods in most cases.
\end{itemize}

\indent
The rest of this paper is organized as follows: Section II introduces related works. Our AE-Rec method is elaborated in Section III. Section IV reports the experimental results on two public datasets. Finally, some conclusions are drawn in Section V.

\section{Related Works}
In this section, we firstly give a brief introduction on the classical RPC model, considering that the RPC model has been widely used in many existing works \cite{2012CARTOSAT,2016BENCHMARK,2016ASP,2017MICMAC} for reconstructing 3D scene structures from satellite images in the remote sensing field. Then, we provide a review on existing 3D reconstruction methods based on optical satellite images. 
\subsection{RPC Model} 
In 1997, Hartley et al. \cite{1997RPC} proposed an algorithm for estimating the parameters of the cubic rational polynomial camera which maps the image points as a rational polynomial function of world coordinates and it could model pushbroom cameras effectively. The frequently-used rational polynomial camera (RPC) model \cite{2004RPC}, which includes 78 coefficients and 10 normalized constants, is defined as:
\begin{equation}
    \begin{matrix}
    r_n=p_1(X_n,Y_n,Z_n)\\
    c_n=p_2(X_n,Y_n,Z_n)
    \end{matrix}, 
\end{equation}
where $r_n$ and $c_n$ are the normalized row and column indices respectively, and $\{X_n,Y_n,Z_n\}$ denote the normalized coordinate values of object points in ground space, $\{p_i\}_{i=1}^2$ are ratios of two cubic polynomials parameterized by 39 coefficients \cite{2004REVIEW}. Although the coefficients in the RPC model have no physical interpretation, it has become a standard practice for satellite image vendors to deliver an RPC model to accompany each satellite image, due to the fact that the RPC model has achieved high accuracies in all stages of the photogrammetry process just as performed by rigorous physical sensor models.\\
\subsection{3D Reconstruction Based on Optical Satellite Images}
In the remote sensing field, most of the existing works for optical-image-based 3D reconstruction methods have been explored by utilizing the RPC model \cite{2012CARTOSAT,2016BENCHMARK,2016ASP,2017MICMAC}. With RPC model parameters provided, these methods usually focus on the minimum configuration of two-view stereo, and the general reconstruction flow involves (1) optimizing parameter estimation residuals between the RPC models via block adjustment \cite{2002BIAS,2004BIAS,2005BIAS,2003BA,2014RECT}, (2) relying on disparity \cite{2012CARTOSAT,2016ASP,2017MICMAC} or optical flow \cite{2016BENCHMARK} to find dense correspondences between views, (3) performing triangulation based on RPC models to reconstruct the 3D scene structure corresponding to optical satellite images \cite{2002RPCTRI,2015MIN}. Franchis et al. \cite{2014S2P} proposed a fully automatic and modular stereo pipeline to produce digital elevation models from optical satellite images. This satellite stereo pipeline, abbreviated as s2p, replaced complicated non-linear bundle adjustment with relative pointing error correction between RPC models and recovers the 3D structure of pairwise satellite images by simple RPC-based elevation iteration. Facciolo et al. \cite{2017S2P} analyzed the influences on two-view reconstruction, and proposed a novel RPC-based method. This method, which heuristically ranked all image pairs in the input set and then selectively merged the independent reconstructions generated by s2p \cite{2014S2P}, won the IARPA Multi-View Stereo 3D Mapping Challenge in 2016.\\
\indent
In addition to the aforementioned RPC-based methods, some works investigated the 3D reconstruction problem for optical satellite images. Fraser et al. \cite{2004AFFINE} employed an affine model to reconstruct the 3D scene structure of entire satellite images, where the parameters of the affine model were calculated directly from the given ground control points. NASA Ames Research Center proposed the NASA Ames Stereo Pipeline, a suite of free and open source automated geodesy and stereogrammetry tools for processing stereo images captured from satellites (around Earth and other planets), where the rigorous physical sensor models used are obtained by querying ephemerides and interpolating camera pose \cite{2018ASP}. Zhang et al. \cite{2019COLMAP} adapted the state-of-the-art reconstruction pipeline COLMAP from the computer vision community to optical satellite image scenarios.

\section{Methodology}
In this section, we propose the AE-Rec framework for 3D Euclidean reconstruction based on optical satellite images. Firstly, we present the design motivation and pipeline of the AE-Rec framework, composed of an affine dense reconstruction stage and an affine-to-Euclidean upgrading stage. Then, we give detailed descriptions on the referred two stages respectively.
\subsection{Design Motivation and Pipeline}
As introduced above, the AE-Rec framework implements a 3D affine dense reconstruction firstly and then an affine-to-Euclidean upgrading, which is motivated by the following two points:
\begin{itemize}
    \item[M1] It is noted that the imaging model of a pushbroom camera is essentially different from that of a plane array camera. However, a pushbroom camera (fixed on a satellite) moves approximately in a straight line in a short time, and the distance between the camera and the ground scene is enormous, hence, the mapping relationship between a small-sized tile in a satellite image and its corresponding 3D local scene could be modelled approximately by an affine camera model \cite{2003MVG}.
    \item[M2] For a given set of satellite images, the calculated 3D affine structure under an affine camera model is up to an affine transformation in comparison to the corresponding Euclidean structure. This affine transformation matrix has 12 degrees of freedom, and it could be uniquely determined by at least 4 GCPs.
\end{itemize}

\indent
Based on the above two points, we design the pipeline of the proposed AE-Rec framework as shown in Fig. 1. As seen from this figure, the proposed method consists of two sequential stages, affine dense reconstruction and affine-to-Euclidean upgrading. Given a set of optical satellite images, they are cropped into a set of small-sized tiles, and these tiles are used to reconstruct an affine dense scene point cloud under an affine camera model based on the aforementioned motivation M1 at the affine dense reconstruction stage. Then at the affine-to-Euclidean stage, a global affine transformation matrix is fitted with at least 4 GCPs for upgrading the obtained 3D affine structure to a Euclidean based on the aforementioned motivation M2. In the following parts, the two stages will be describe in detail.
\begin{figure*}[htbp]
    \centering
    \includegraphics[scale=0.8]{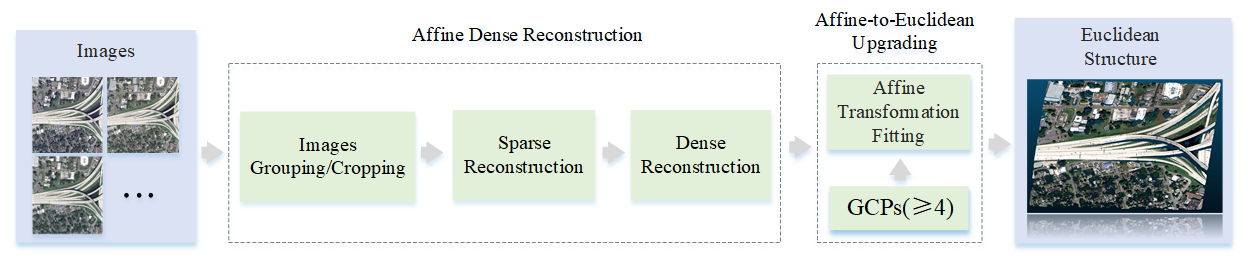}
    \caption{AE-Rec framework.}
    \label{AE-Rec framework.}
\end{figure*}
\subsection{Affine Dense Reconstruction Stage}
In the affine dense reconstruction stage, a dense 3D affine reconstruction approach is explored for recovering the 3D scene structure from an input set of optical satellite images in an affine space under an incremental reconstruction strategy. The explored approach consists of three main steps as:
\begin{itemize}
    \item[S1] Image Grouping and Cropping: Considering that an arbitrary pair of images, which are captured for a same region with a relatively long time interval respectively, generally have much different textures for hampering accurate point matching, the input set of satellite images are firstly grouped into several subsets under an introduced image grouping criterion. Then, the grouped large-sized satellite images are cropped into numerous small-sized tiles with partial overlapping regions, so that these tiles could be effectively modeled by an affine camera model. 
    \item[S2] Sparse Reconstruction: For the tiles belong to each grouped subset of satellite images at Step S1, a high-quality point feature extraction/matching algorithm (here, SIFT \cite{2004SIFT} is straightforwardly used) is implemented for obtaining sparse point correspondences, and then a sparse affine reconstruction is implemented with these obtained point correspondences for obtaining a sparse affine scene structure and a set of affine camera motion matrices.
    \item[S3] Dense Reconstruction: For each pair of tiles (belonging to an arbitrary grouped subsets of satellite images at Step S1) with overlapping regions, an affine stereo rectification is implemented, and then a stereo matching algorithm \cite{2015MGM} is performed on these rectified tiles for obtaining dense feature correspondences. Both the obtained dense correspondences and affine camera motion matrices at Step S2 are jointly used for reconstructing a dense 3D affine structure.
\end{itemize}

\indent
Here, it has to be explained the reason of why the proposed affine dense reconstruction approach implements sparse reconstruction before dense reconstruction, rather than only implementing dense reconstruction: Generally compared with state-of-the-art sparse matching algorithms, the existing dense matching algorithms have a much larger number of point correspondences, but a lower matching accuracy. If the dense reconstruction is straightforwardly implemented by utilizing dense point correspondences (obtained by a dense matching algorithm) without a prior sparse reconstruction, the estimated global camera motion matrices would be relatively poorer due to the obtained point correspondences with a low matching accuracy, resulting in a relatively lower reconstruction accuracy. If a sparse reconstruction is implemented by utilizing sparse point correspondences (obtained by a sparse matching algorithm) before dense reconstruction, the estimated global camera motion matrices would be relatively better, resulting in a relatively better reconstruction accuracy.\\
\indent
In the following parts, we will give a detailed introduction to the above three steps:
\subsubsection{Image Grouping and Cropping}
As indicated above, an arbitrary pair of images that are captured for a same region with a relatively long time interval respectively, generally have much different textures for hampering accurate point matching. Hence, in input set of satellite images which are captured during a long time range, we introduce the following image grouping criterion to group the input set of images into several subsets: Images with capture time intervals less than 15 days are treated as the same grouped subset. In subsequent steps, we process each subset of optical satellite images independently.\\
\indent
Then, the large-sized satellite images in each subset are cropped into tiles with a same size $d\times{d}$ (d is much smaller than the size of the input images), subject to the condition that each pair of neighboring tiles have an overlapping regions with a size of $\frac{d}{3}\times{d}$ (or $d\times\frac{d}{3}$), as shown in Fig. 2. Since these tiles are of a smaller size, they could be modeled approximately by an affine camera model. In addition, the overlapping regions belonging to neighboring tiles guarantee that the calculated local point clouds in the following sparse reconstruction step could be registered into a unified affine coordinate system. \\
\begin{figure}[htbp]
	\centering
	\includegraphics[width=0.45\textwidth]{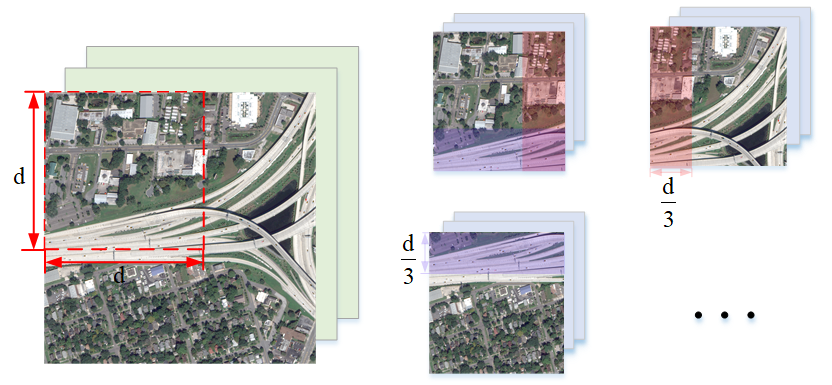}
	\caption{Images cropping sample.}
	\label{Fig.2.} 
\end{figure} 
\subsubsection{Sparse Reconstruction}
Based on the obtained tiles above, a sparse reconstruction is implemented to estimate a set of global affine camera motion matrices in an incremental reconstruction strategy (as shown in Fig. 3), where an affine reconstruction of an selected initial pair of tiles is firstly built and then the other tiles are added iteratively to the reconstruction one at a time. This sparse reconstruction is built on the scene graph resulting from feature matching, and includes three main modules: affine reconstruction initialization, new tile addition, and local optimization.\\
\begin{figure*}[htbp]
    \centering
    \includegraphics[scale=1]{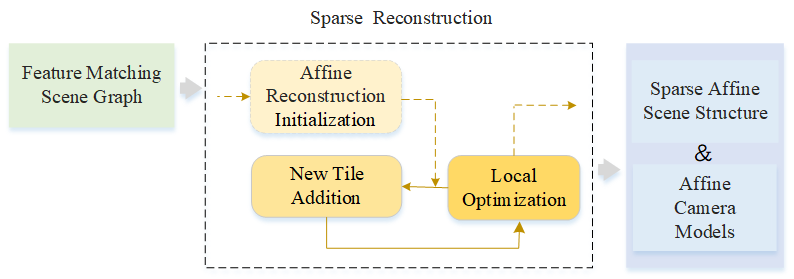}
    \caption{Sparse reconstruction incremental strategy.}
    \label{Sparse reconstruction incremental strategy.}
\end{figure*}
\indent
\textbf{Affine Reconstruction Initialization:} Since all optical satellite images used in the affine dense reconstruction are captured over the same site, in order to obtain more robust and accurate reconstruction results, we select the initial image pair first, and then choose the initial tile pair used for the affine reconstruction initialization in the initial image pair. The sparser the matching points indicate that the angle between the two optical satellite images is larger, and such an image pair tends to obtain more accurate reconstruction results. Hence, we select the image pair with the sparsest SIFT matching as the initial image pair. In the initial image pair, we further separately count the number of matches in all small-sized tile pairs, and choose the small-sized tile pair with the largest SIFT matching number as the initial tile pair, which will ensure the quantity of reconstructed 3D points in the affine reconstruction initialization so that sufficient visible points are available for subsequent new tile additions.\\
\indent
With the aforementioned selected initial tile pair, we implement the affine reconstruction initialization using the factorization affine reconstruction algorithm \cite{1992Fact}. Let Tile $i$ and Tile $j$ denote an arbitrary pair of tiles, which have a set of $n(\geq4)$ correspondences $\{(\textbf{x}_i^1, \textbf{x}_j^1),\cdots,(\textbf{x}_i^k, \textbf{x}_j^k),\cdots, (\textbf{x}_i^n, \textbf{x}_j^n)\}$. $\textbf{x}_i^k$ and $\textbf{x}_j^k$ $\in$ $R^2$ ($k=1,\cdots,n$) represent the inhomogeneous coordinates of the $k$-th matching point in Tile $i$ and Tile $j$ respectively, and the inhomogeneous coordinates of their corresponding 3D point in the affine space is denoted as $\textbf{X}_{ij}^k$. Let $A_i=[M_i,\textbf{t}_i]$ and $A_j=[M_j,\textbf{t}_j]$ denote the corresponding affine camera models to Tile $i$ and Tile $j$ respectively. As defined in \cite{2003MVG}, the mapping relationship between $\textbf{X}_{ij}^k$ and its corresponding image point $\textbf{x}_i^k$ in Tile $i$ ($\textbf{x}_j^k$ in Tile $j$) under the affine camera model $A_i$ (also $A_j$) is formulate as:
\begin{equation}
    \textbf{x}_i^k=A_i[\textbf{X}_{ij}^k;1]=M_i{\textbf{X}_{ij}^k}+\textbf{t}_i.
\end{equation}
Given a set of $n$ correspondences between Tile $i$ and Tile $j$, $\textbf{t}_i$ in $A_i$ could be omitted by centralized the image points $\{\textbf{x}_i^k\}_{k=1}^{k=n}$ into $\{\bar{\textbf{x}}_i^k\}_{k=1}^{k=n}$ (i.e. $\{\bar{\textbf{x}}_i^k\}=\{\textbf{x}_i^k\}-1/n\sum\textbf{x}_i^k)$, while $\textbf{t}_j$ could be omitted by centralized the image points $\{\bar{\textbf{x}}_j^k\}_{k=1}^{k=n}$. Accordingly, the following constraint is obtained:
\begin{equation}
    [\bar{\textbf{x}}_i^k;\bar{\textbf{x}}_j^k]=[M_i;M_j]\textbf{X}_{ij}^k.
\end{equation}
As noted from (3), once $n$ is larger than 4, the affine space 3D points coordinates $\{\textbf{X}_{ij}^k\}$ and the matrices $\{M_i,M_j\}$ could be obtained by minimizing the following objective functions:
\begin{equation}
    \min_{M_i,M_j,\textbf{X}_{ij}^k}\sum_k||[\bar{\textbf{x}}_i^k;\bar{\textbf{x}}_j^k]-[M_i;M_j]\textbf{X}_{ij}^k||_F^2.
\end{equation}
Here, a closed-form solution to the above minimization problem is simply obtained by SVD as indicated by matrix optimization theory \cite{1992Fact}: if the SVD of matrix $W=UDV^\top$, where W is a $4\times{n}$ matrix formed by a set of $n$ centralized correspondences, 
\begin{equation}
    W=
    \left[
    \begin{matrix}
    \bar{\textbf{x}}_i^1 & \cdots & \bar{\textbf{x}}_i^k & \cdots & \bar{\textbf{x}}_i^n\\
    \bar{\textbf{x}}_j^1 & \cdots & \bar{\textbf{x}}_j^k & \cdots & \bar{\textbf{x}}_j^n
    \end{matrix}
    \right].
\end{equation}
then the product of the first three columns of $U$ and the third-order submatrix in the upper left corner of $D$ is $[M_i;M_j]$, and the 3D points coordinates $\textbf{X}_{ij}^k$ in affine space can be extracted from the first three columns of $V$. For more detail, refer to Algorithm 1.
\begin{algorithm}[htb]
\caption{ Affine Reconstruction Initialization \cite{2003MVG}. }
\label{alg:affine reconstruction}
\begin{algorithmic}[1] 
\REQUIRE ~~\\ 
    Initial tiles pair correspondences $(\textbf{x}^k_i,\textbf{x}^k_j)$, $k=1,\cdots,n$;
\ENSURE ~~\\ 
    Affine camera matrix $A_i=\left[ \begin{matrix} M_i & \textbf{t}_i \end{matrix} \right]$, $Aj=\left[ \begin{matrix} M_j & \textbf{t}_j \end{matrix} \right]$, and 3D points $\{\textbf{X}_{ij}^k\}$;
    \STATE Compute translation vectors $\textbf{t}_i$ and $\textbf{t}_j$, namely the centroid of points in Tile $i$ (or Tile $j$),
    $$
    \textbf{t}_{i,j}=\frac{1}{n}\sum_k\textbf{x}_{i,j}^k.
    $$
    \STATE Centralized image point coordinates:
    $$
    \{\bar{\textbf{x}}_{i,j}^k\}=\{\textbf{x}_{i,j}^k\}-\frac{1}{n}\sum_k\textbf{x}_{i,j}^k,
    $$
    \STATE Construct the $4\times{n}$ matrix $W$ from the centralized correspondence, and comput its SVD 
    $$
    W=UDV^\top.
    $$
    \STATE Then the matrices $\{M_i,M_j\}$ are obtained from the product of first three columns of $U$ and the first three singular values in $D$:
    $$
    \left[ \begin{array}{rcl} M_i \\ M_j \end{array} \right] = \left[ \begin{matrix} \sigma_1\textbf{u}_1 & \sigma_2\textbf{u}_2 & \sigma_3\textbf{u}_3 \end{matrix} \right].
    $$
    and $\{\textbf{X}_{ij}^k\}$ are read from the first three columns of $V$
    $$
    \left[ \begin{matrix} \textbf{X}_{ij}^1 & \cdots & \textbf{X}_{ij}^k & \cdots & \textbf{X}_{ij}^n \end{matrix} \right] = \left[ \begin{matrix} \textbf{v}_1 & \textbf{v}_2 & \textbf{v}_3 \end{matrix} \right]^\top.
    $$
\end{algorithmic}
\end{algorithm}

\indent
\textbf{New Tile Addition:}
Starting from the affine reconstruction initialization, the new tile addition is implemented by running two separate procedures of new tile registration and affine triangulation, alternatively. A new tile can be registered to the current affine 3D structure by using feature correspondences to reconstructed points in already registered tiles (2D-3D correspondences), thereby extending the set of reconstructed 3D points in the affine space through affine triangulation to increase scene coverage. The mapping realtionship of affine camera model in (2) can be described in a more general form, 
\begin{equation}
    \begin{matrix}
    r_i^k=A_i^1X^k+A_i^2Y^k+A_i^3Z^k+A_i^4\\
    c_i^k=A_i^5X^k+A_i^6Y^k+A_i^7Z^k+A_i^8
    \end{matrix},
\end{equation}
where $r_i^k$ and $c_i^k$ are the row and column indices of the new tile image space point $\textbf{x}_i^k$ in a 2D-3D correspondence respectively, and $(X^k,Y^k,Z^k)$ represents the coordinates of its corresponding 3D point in the current affine structure, $A^1$ to $A^8$ are the eight affine camera model parameters of the new tile. As can be seen from (6), these eight parameters per new tile require a set of $n(\geq4)$ 2D-3D correspondences to be linearly solved. Due to the fact that 2D-3D correspondences are often contaminated with outliers, the RANSAC (random sample consistency) \cite{1981RANSAC} algorithm is adopted for the robust estimation of the affine camera model parameters. Furthermore, we design a selection strategy to add new tile: there is no doubt that the tiles in the initial image pair are registered first according to the nearest-first principle, so that a global affine reconstruction structure about the initial image pair can be obtained. On this basis, we determine the order of adding new tiles according to the number of correspondences with the intial image pair as following the selection criteria for the initial tile pair.\\
\indent
A new 3D point can be affine triangulated and added to the global affine reconstruction points set as soon as a new tile covering the new scene is successfully registered. We implement the multi-view affine triangulation via the inverse of the affine camera matrix, 
\begin{equation}
    [\textbf{X}^k;1]=[A_1;\cdots;A_i;\cdots;A_n]^{-1}[\textbf{x}_1^k;\cdots;\textbf{x}_i^k;\cdots;\textbf{x}_n^k],
\end{equation}
where $A_i$ denotes the affine camera model corresponding to each tile in the multi-view, and $\textbf{x}_i^k$ is the correspondence point which come from each tile, $\textbf{X}^k$ is the coordinate of the 3D point in the affine space obtained by affine triangulation. However, we perform triangulation only on the corresponding tiles from different images, to prevent redundant 3D points being reconstructed. Hence, we form feature tracks by concatenating the correspondences across corresponding tiles, and perform multi-view affine triangulation with these feature tracks.\\
\indent
\textbf{Local Optimization:}  Without further refinement methods, the correspondences with noise will quickly propagate large errors during the iterations of tile registration and affine triangulation, resulting in the affine structure drifting to a non-recoverable state. We employ BA (Bundle Adjustment) \cite{1999BA} for the joint non-linear refinement of camera parameters $A_i$ and 3D point parameters $\textbf{X}_i^k$ to minimize the affine reprojection error, 
\begin{equation}
\min_{A_i,\textbf{X}_i^k}\sum_i\sum_k||\pi(A_i,\textbf{X}_i^k)-\textbf{x}_i^k||_\gamma,
\end{equation}
using a function $\pi$ that projects 3D points in the affine space to image space and $||\cdot||_\gamma$ is a robust norm to potentially down-weight outlies. The optimization problems of bundle adjustments in affine reconstruction are solved with the Levenberg-Marquardt \cite{1944LM} algorithm implemented using Ceres \cite{ceres-solver}.
\subsubsection{Dense Reconstruction}
In order to obtain dense point correspondences for reconstructing a dense affine structure, affine stereo rectification and stereo matching are performed sequentially on each pair of tiles with a sufficient number of sparse point correspondences. It is noted that the typical stereo rectification algorithms in literature \cite{1985STEREO,2000RECT} are not available to handle large-sized satellite images which are generally captured by a pushbroom camera, hence, we introduce a local affine stereo rectification algorithm for re-projecting each pair of small-sized tiles to a new common plane parallel to the line between optical centers, which is outlined in Algorithm 2: given at least 4 correspondences from a tile pair, we estimate the affine fundamental matrix $F_A$ between tiles in this paire employing the Gold Standard algorithm \cite{2003MVG}, and then two rectifying affine transformations are extracted from $F_A$ \cite{1999ZZY}, for rectifying the respective tiles. Once the pair of rectified tiles are obtained, MGM \cite{2015MGM} is implemented on them for stereo matching, resulting in a set of dense point correspondences.
\begin{algorithm}[htb]
\caption{ Affine stereo rectification \cite{2003MVG, 1999ZZY}. }
\label{alg:stereo-rectification}
\begin{algorithmic}[1] 
\REQUIRE ~~\\ 
    Tile $i$ and Tile $j$, and their correspondences;
\ENSURE ~~\\ 
    Rectified Tile $i_{rect}$ and Tile $j_{rect}$;
    \STATE Estimate $F_A$ by the Gold Standard algorithm;
    \STATE Extract $H_i$ and $H_j$ from $F_A$;
    \STATE Rectify Tile $i$ (Tile $j$) to Tile $i_{rect}$ (Tile $j_{rect}$) with $H_i$ ($H_j$) respectively.
\end{algorithmic}
\end{algorithm}

\indent
After obtaining the dense correspondences and the global affine camera matrices, we use both of them jointly to reconstruct the dense 3D affine structure. For multi-view dense affine reconstruction, we transit the dense correspondences across multi-views to obtain dense matches tracks. Considering outliers and redundancies in the dense correspondence, a photometric consistency constraint is imposed on all dense matches tracks, which allows optimization to split tracks connected by weak correspondences, providing robustness to mismatches. Once all dense matches tracks are determined, we reconstruct the dense 3D point cloud in the affine space by multi-view affine triangulation in (7). Eventually, the 3D point coordinates of the dense affine reconstruction are optimized with BA to improve the geometric consistency of the global affine structure.
\subsection{Euclidean Upgrading}
It is noted that the obtained 3D scene structure at the above subsection is an affine structure, which is up to an affine transformation $H_A \in R^{4\times4}$ (under the homogeneous coordinate system) in comparison to the corresponding Euclidean structure:
\begin{equation}
    H_A\textbf{X}_A=\textbf{X}_E,
\end{equation}
where $\textbf{X}_A$ is an reconstructed 3D scene point in the affine coordinate system, and $\textbf{X}_E$ is the corresponding ground truth 3D point in the Euclidean coordinate system. Since the last row of an arbitrary affine transformation matrix with a size of $4\times4$ is $(0,0,0,1)$, $H_A$ has indeed 12 degrees of freedom, and it could be uniquely determined by at least 4 non-coplanar GCPs (with their corresponding image points) and their corresponding affine points which could be straightforwardly found according to the obtained mapping relationship between the tile image space and the affine space at Subsection B. In detail, assuming that $N(\geq4)$ couples of GCPs in the Euclidean space and their corresponding points in the affine space are given. $H_A$ is calculated by solving the following least-squares problem:
\begin{equation}
    \min_{H_A}\sum_j||H_A\textbf{X}_A^j-\textbf{X}_E^j||_2^2.
\end{equation}
\\
\indent
In addition, when $N$ is much larger than 4, RANSAC \cite{1981RANSAC} could also be employed for achieving a more robust estimation by automatically removing possible outliers form the given couples of GCPs and their affine points.
\subsection{Optional Post-processing}
Since the input set of optical satellite images is grouped into several subsets (in Section B), our proposed AE-Rec framework reconstructs an independent Euclidean structure for each grouped subset. Although the different textures among the images in different subsets make the appearance changes among the Euclidean structures corresponding to these subsets. However, all images in these subsets cover the same geographic area, and if a more complete reconstruction is pursued, we perform optional post-processing on the reconstructed Euclidean structures corresponding to these subsets, i.e. utilizing the ICP (Iterative Closet Point) \cite{1987ICP} algorithm to align and fuse several reconstructed Euclidean structures corresponding to the same geographic region.

\section{Experiments}
In this section, we evaluate the proposed AE-Rec method and three state-of-the-art methods, including JHU/APL method \cite{2016BENCHMARK}, S2P method \cite{2017S2P}, COLMAP method \cite{2019COLMAP}. Firstly, we introduce two benchmark datasets and evaluation metrics. Then, we investigate the influence of image cropping size $d$ on the reconstruction performance. Finally, the comparative evaluation on the proposed method and three state-of-the-art methods is implemented.
\subsection{Datasets and Evaluation Metrics}
To evaluate the effectiveness of our method, we conduct extensive experiments on the following two benchmark datasets:
\begin{enumerate}
	\item MVS3DM Dataset\cite{2016BENCHMARK}: MVS3DM dataset is provided by IARPA (Intelligence Advanced Research Projects Activity), which contains fifty WorldView-3 panchromatic, visible, and near infrared (VNIR) images, captured from November 2014 to January 2016 near San Fernando, Argentina. The ground truth DEM for evaluation was collected by airborne radar which ground sample distance (GSD) is approximately 30cm. Johns Hopkins University has preprocessed the dataset and identified eight reconstructable sites based on image overlap regions and LiDAR ranges.
	\item DFC2019 Dataset\cite{2019DFC}: DFC2019 dataset comes from the IEEE Geoscience and Remote Sensing (GRSS) Data Fusion Competition 2019. It consists of 2783 images from 106 observation sites corresponding to Jack Wilson, Florida, and Omaha, Nebraska. These WorldView-3 images were also collected from 2014 to 2016. The size is $2048\times2048$ for each RGB images, while the ground truth DSM size is $512\times512$.
\end{enumerate}

\indent
As done in \cite{2017METRIC}, in order to evaluated the performances of all the comparative methods, we project the height values of the reconstructed 3D point clouds onto a normalized geographic grid by nearest-neighbor interpolation, with geographic grid units equal to the ground sample distance of ground truth DEMs in each dataset. The DEM generated by each comparative method are aligned to the ground-truth DEM, and the
height values are compared pixel-wise to generate the height error maps. Then, the following two metrics are used for evaluation:
\begin{itemize}
    \item Median height error: the median value in the height error map;
    \item Completeness: the percentage of points with height error less than a threshold of 1m in the height error.
\end{itemize}
\begin{table}
	\centering
	\caption{COMPLETENESS($\%$) AND ACCURACY($m$) OF AE-REC WITH DIFFERENT IMAGE CROPPING SIZES}
	\begin{tabular}{cccc}
		\toprule
		$d$ & CP(\%) & ME($m$)\\
		\midrule
		 300 & 71.0 & 0.269\\
		 500 & 72.5 & 0.252\\
		 800 & 72.8 & 0.244\\
		1000 & 72.9 & 0.239\\
		1500 & 73.0 & 0.242\\
		2000 & 72.2 & 0.240\\
		\bottomrule
	\end{tabular}
\end{table}
\subsection{Influence of Image Cropping Size}
We evaluate the influence of image cropping size $d$ on the reconstruction performance by reconstructing two-views satellite images through the proposed method with $d=\{300, 500, 800, 1000, 1500, 2000\}$ respectively. The corresponding results are reported in Table 1. As seen from this table, the proposed method with $d=1000$ achieve a trade-off among all the evaluation metrics, hence, this image cropping size $d$ is always set to $1000$ in all of our experiments. It is also noted that when $d$ ranges from $800$ to $2000$, the corresponding results are quite close, demonstrating that the performance of the propose method is not sensitive to this image cropping size $d$ in this range.
\subsection{Comparison on Benchmark Data}
Now, we evaluate our proposed method and compare it comprehensively with the experiment results of three state-of-the-art methods. Since both datasets were captured with a relatively long time interval, the optical satellite image set corresponding each site in both datasets can be grouped into several image subsets. For the sake of fairness, the comparison experiments will be conducted in two parts. First, we evaluate the accuracy and completeness of the aforementioned methods using the selected subsets of images corresponding to each site in both datasets. And then, the reconstruction performance and run-time of these methods are evaluated in the complete image set of these sites.
\subsubsection{For the selected image subset}
Here, we report the evaluation results for 15 sites in the aforementioned two datasets. Among them, 8 sites are from the MVS3DM dataset and the image subsets of these sites are provided by Johns Hopkins University (containing 3-4 images). The other 7 sites are from the DFC2019 dataset, and the image subset corresponding to each site is the largest subset selected according to the image grouping method described above (containing 5-8 images). The evaluation results for the two datasets are presented in Table 2 and Table 3, respectively, and we observe that the proposed AE-Rec framework outperforms the other three state-of-the-art methods. It ensure that the reconstructed DSMs obtain the highest completeness while maintaining the lowest median height error among all methods, which is consistent with the visualization of the height error maps in Fig. 4 and Fig. 5.\\
\begin{table}[!tb]
\footnotesize 
\caption{COMPARISON OF QUANTITATIVE RESULTS ON SELECTED SUBSETS OF MVS3DM DATASET.}
\label{tab:Vaihingen_result}
\centering
\begin{tabular*}{\linewidth}{@{\extracolsep{\fill}}@{\quad}l@{\extracolsep{\fill}}@{\quad}c@{\extracolsep{\fill}}@{\quad}c@{\extracolsep{\fill}}@{\quad}c@{\extracolsep{\fill}}@{\quad}c@{\extracolsep{\fill}}@{\quad}c@{\extracolsep{\fill}}@{\quad}c}
\hline
MVS3DM    &   Method  &   CP(\%)    &   ME($m$)  \\
\hline
\multirow{4}{*}{Site1} & JHU/APL\cite{2016BENCHMARK}    & 64.0 & 0.577 \\
                       & COLMAP\cite{2019COLMAP}
  & 61.7 & 0.276 \\
                       & S2P\cite{2017METRIC} 
  & 72.4 & 0.319 \\
                       & Ours 
  & \textbf{74.7} & \textbf{0.259} \\	

\hline
\multirow{4}{*}{Site2} & JHU/APL\cite{2016BENCHMARK}    & 62.3 & 0.597 \\
                       & COLMAP\cite{2019COLMAP}
  & 63.1 & 0.302 \\
                       & S2P\cite{2017METRIC} 
  & 67.1 & 0.340 \\
                       & Ours 
  & \textbf{70.9} & \textbf{0.302} \\	
\hline
\multirow{4}{*}{Site3} & JHU/APL\cite{2016BENCHMARK}    & 52.9 & 0.614 \\
                       & COLMAP\cite{2019COLMAP}
  & 31.9 & 0.364 \\
                       & S2P\cite{2017METRIC} 
  & 63.6 & 0.392 \\
                       & Ours 
  & \textbf{65.7} & \textbf{0.345} \\	
\hline
\multirow{4}{*}{Site4} & JHU/APL\cite{2016BENCHMARK}    & 39.7 & 1.346 \\
                       & COLMAP\cite{2019COLMAP}
  & 10.9 & 0.632 \\
                       & S2P\cite{2017METRIC} 
  & 40.6 & 0.380 \\
                       & Ours 
  & \textbf{41.8} & \textbf{0.363} \\	
\hline
\multirow{4}{*}{Site5} & JHU/APL\cite{2016BENCHMARK}    & 70.3 & 0.389 \\
                       & COLMAP\cite{2019COLMAP}
  & 62.4 & 0.312 \\
                       & S2P\cite{2017METRIC} 
  & 67.2 & 0.369 \\
                       & Ours 
  & \textbf{71.4} & \textbf{0.272} \\
\hline
\multirow{4}{*}{Site6} & JHU/APL\cite{2016BENCHMARK}    & 66.6 & 0.442 \\
                       & COLMAP\cite{2019COLMAP}
  & 58.9 & 0.323 \\
                       & S2P\cite{2017METRIC} 
  & 64.6 & 0.381 \\
                       & Ours 
  & \textbf{67.8} & \textbf{0.283} \\
\hline
\multirow{4}{*}{Site7} & JHU/APL\cite{2016BENCHMARK}    & 49.9 & 0.916 \\
                       & COLMAP\cite{2019COLMAP}
  & 33.9 & 0.373 \\
                       & S2P\cite{2017METRIC} 
  & 48.5 & 0.317 \\
                       & Ours 
  & \textbf{50.1} & \textbf{0.298} \\	
\hline
\multirow{4}{*}{Site8} & JHU/APL\cite{2016BENCHMARK}    & 54.3 & 0.601 \\
                       & COLMAP\cite{2019COLMAP}
  & 20.0 & 0.452 \\
                       & S2P\cite{2017METRIC} 
  & 51.0 & 0.322 \\
                       & Ours 
  & \textbf{51.7} & \textbf{0.320} \\	
\hline
\end{tabular*}
\end{table}

\begin{table}[!tb]
\footnotesize 
\caption{COMPARISON OF QUANTITATIVE RESULTS ON SELECTED SUBSETS OF DFC2019 DATASET.}
\label{tab:Vaihingen_result}
\centering
\begin{tabular*}{\linewidth}{@{\extracolsep{\fill}}@{\quad}l@{\extracolsep{\fill}}@{\quad}c@{\extracolsep{\fill}}@{\quad}c@{\extracolsep{\fill}}@{\quad}c@{\extracolsep{\fill}}@{\quad}c@{\extracolsep{\fill}}@{\quad}c@{\extracolsep{\fill}}@{\quad}c}
\hline
DFC2019    &   Method  &   CP(\%)    &   ME($m$)  \\
\hline
\multirow{4}{*}{Site1} & JHU/APL\cite{2016BENCHMARK}    & 2.0 & 14.070 \\
                       & COLMAP\cite{2019COLMAP}
  & 51.3 & 0.443 \\
                       & S2P\cite{2017METRIC} 
  & 76.8 & 0.267 \\
                       & Ours 
  & \textbf{82.1} & \textbf{0.235} \\	

\hline
\multirow{4}{*}{Site2} & JHU/APL\cite{2016BENCHMARK}    & 64.9 & 0.603 \\
                       & COLMAP\cite{2019COLMAP}
  & 45.1 & 0.344 \\
                       & S2P\cite{2017METRIC} 
  & 68.2 & 0.462 \\
                       & Ours 
  & \textbf{72.1} & \textbf{0.337} \\	
\hline
\multirow{4}{*}{Site3} & JHU/APL\cite{2016BENCHMARK}    & 55.9 & 0.814 \\
                       & COLMAP\cite{2019COLMAP}
  & 41.4 & 0.451 \\
                       & S2P\cite{2017METRIC} 
  & 65.7 & 0.223 \\
                       & Ours 
  & \textbf{67.7} & \textbf{0.186} \\	
\hline
\multirow{4}{*}{Site4} & JHU/APL\cite{2016BENCHMARK}    & 69.8 & 0.450 \\
                       & COLMAP\cite{2019COLMAP}
  & 57.5 & 0.349 \\
                       & S2P\cite{2017METRIC} 
  & 72.2 & 0.330 \\
                       & Ours 
  & \textbf{73.9} & \textbf{0.216} \\	
\hline
\multirow{4}{*}{Site5} & JHU/APL\cite{2016BENCHMARK}    & 72.2 & 0.344 \\
                       & COLMAP\cite{2019COLMAP}
  & 33.4 & 0.224 \\
                       & S2P\cite{2017METRIC} 
  & 77.8 & 0.162 \\
                       & Ours 
  & \textbf{78.4} & \textbf{0.141} \\
\hline
\multirow{4}{*}{Site6} & JHU/APL\cite{2016BENCHMARK}    & 72.7 & 0.279 \\
                       & COLMAP\cite{2019COLMAP}
  & 52.5 & 0.245 \\
                       & S2P\cite{2017METRIC} 
  & 80.4 & 0.183 \\
                       & Ours 
  & \textbf{82.6} & \textbf{0.148} \\
\hline
\multirow{4}{*}{Site7} & JHU/APL\cite{2016BENCHMARK}    & 93.0 & 0.129 \\
                       & COLMAP\cite{2019COLMAP}
  & 73.9 & 0.112 \\
                       & S2P\cite{2017METRIC} 
  & 96.4 & 0.076 \\
                       & Ours 
  & \textbf{96.5} & \textbf{0.072} \\	
\hline
\end{tabular*}
\end{table}

\begin{figure}[htbp]
    \centering
    \quad
    \subfigure[Ground-Truth \& COLOR BAR]{
        \includegraphics[scale=0.05]{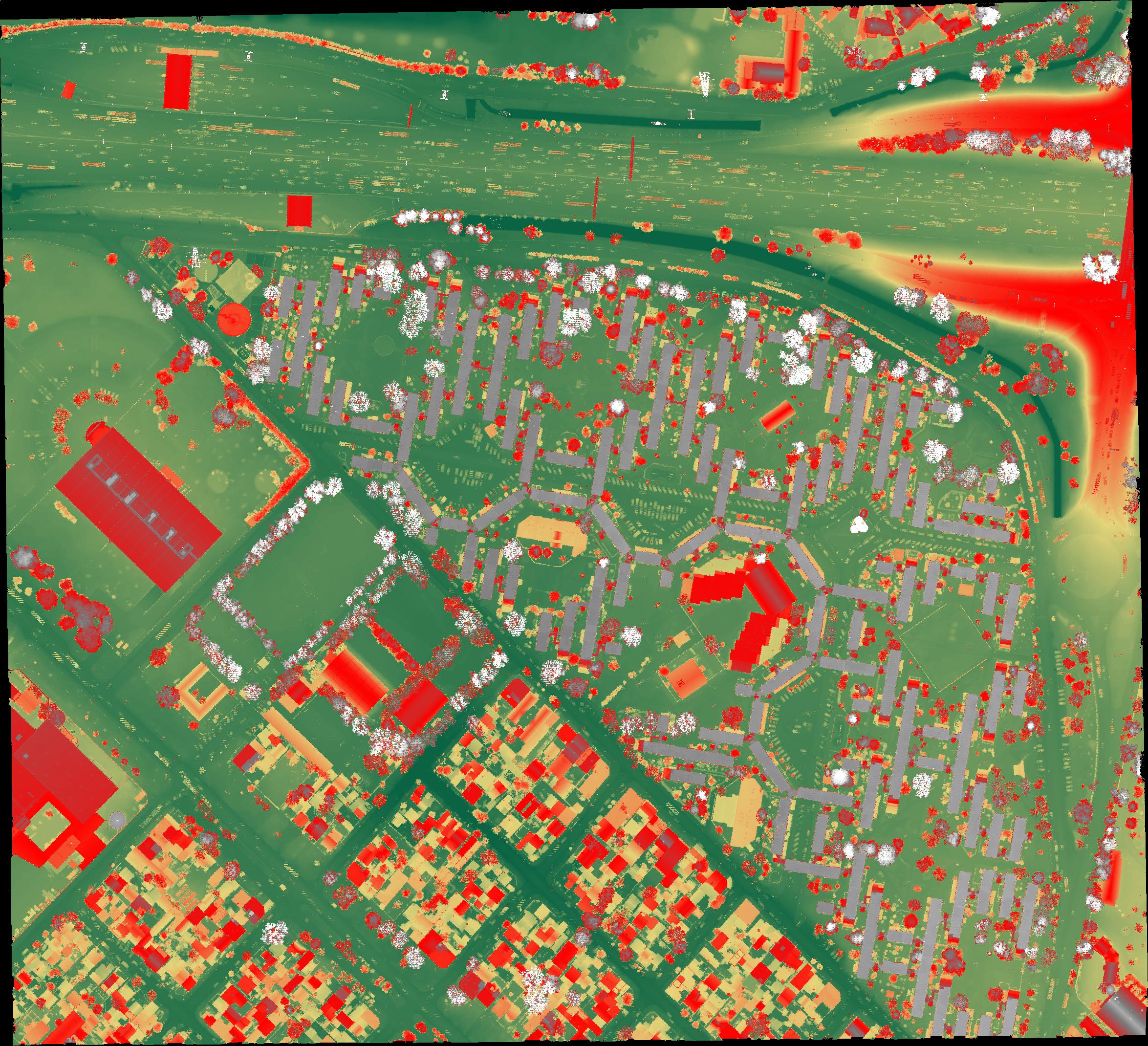} 
        \includegraphics[scale=0.3]{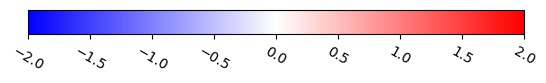} 
    }
    \quad
    \subfigure[JHU/APL\cite{2016BENCHMARK}]{
        \includegraphics[scale=0.05]{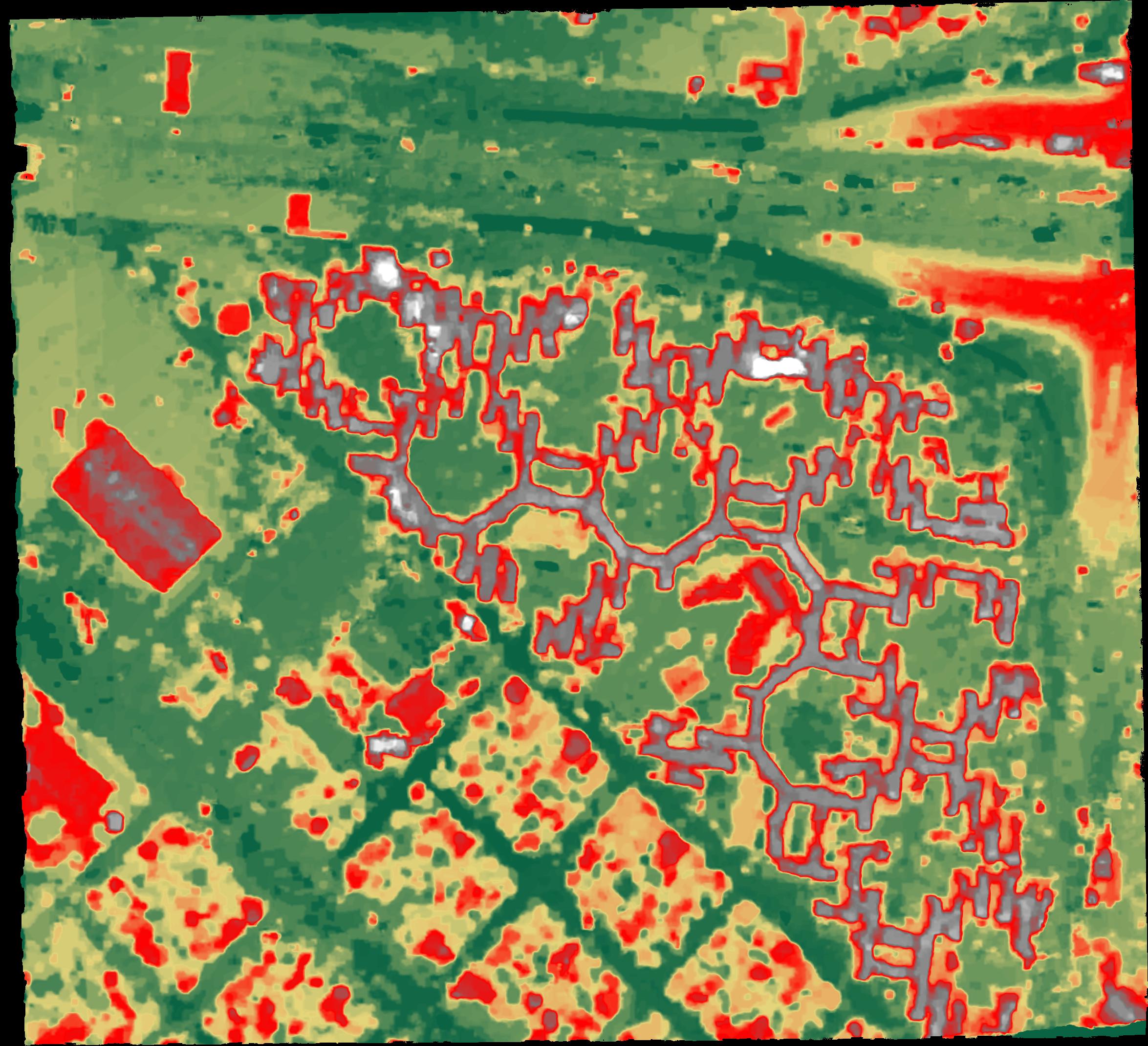} 
        \includegraphics[scale=0.05]{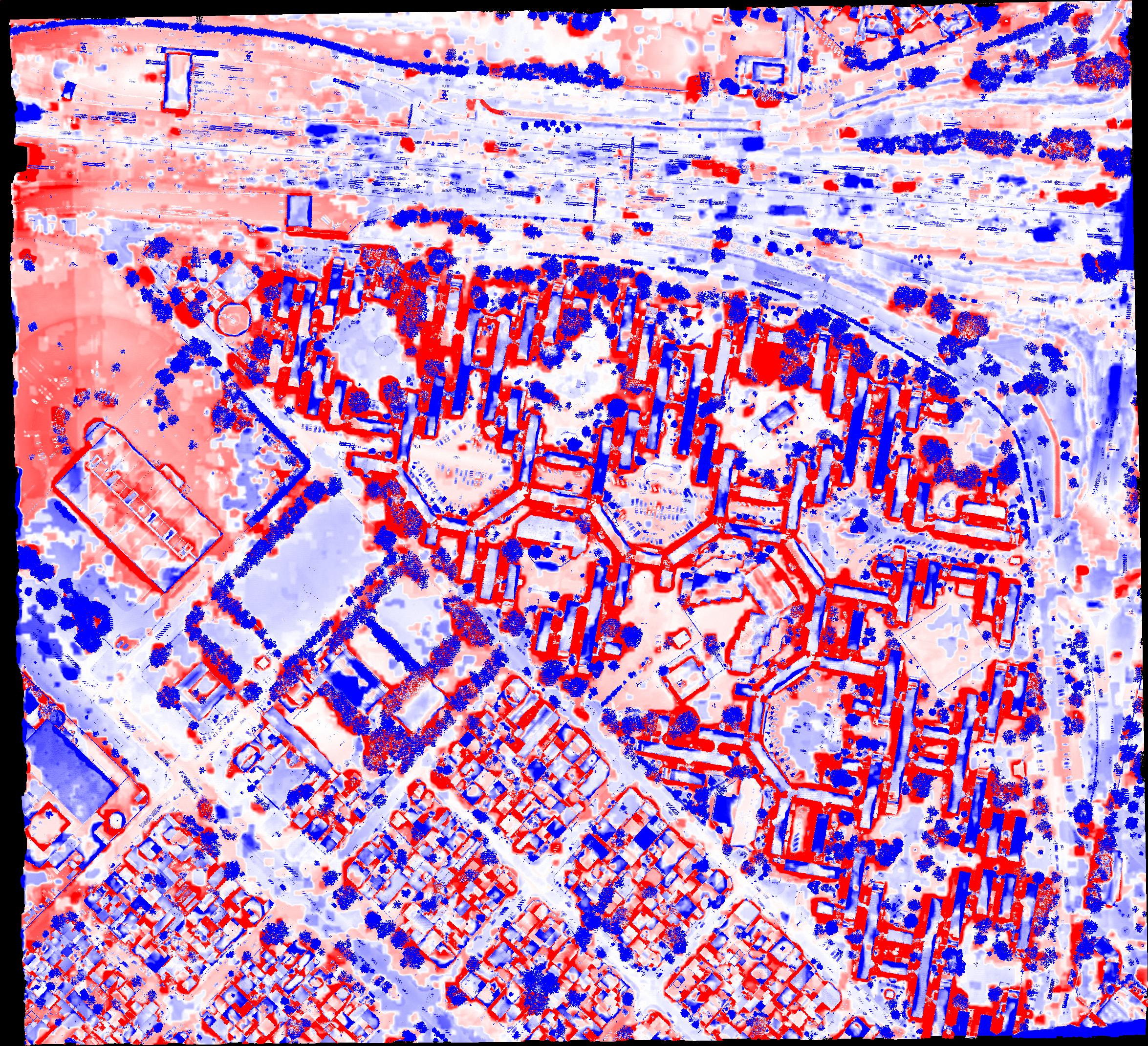} 
    }
    \quad
    \subfigure[COLMAP\cite{2019COLMAP}]{
        \includegraphics[scale=0.05]{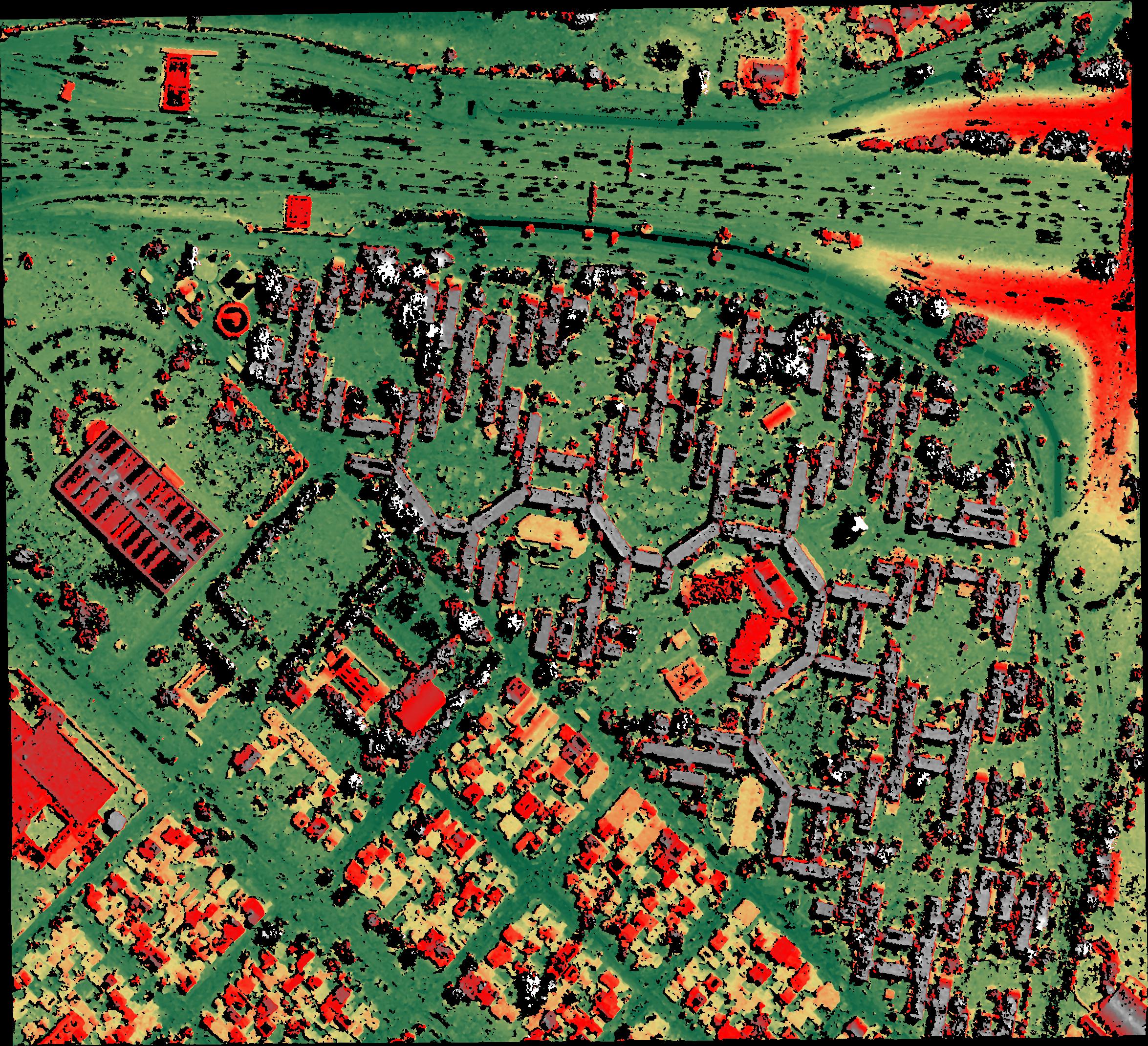}
        \includegraphics[scale=0.05]{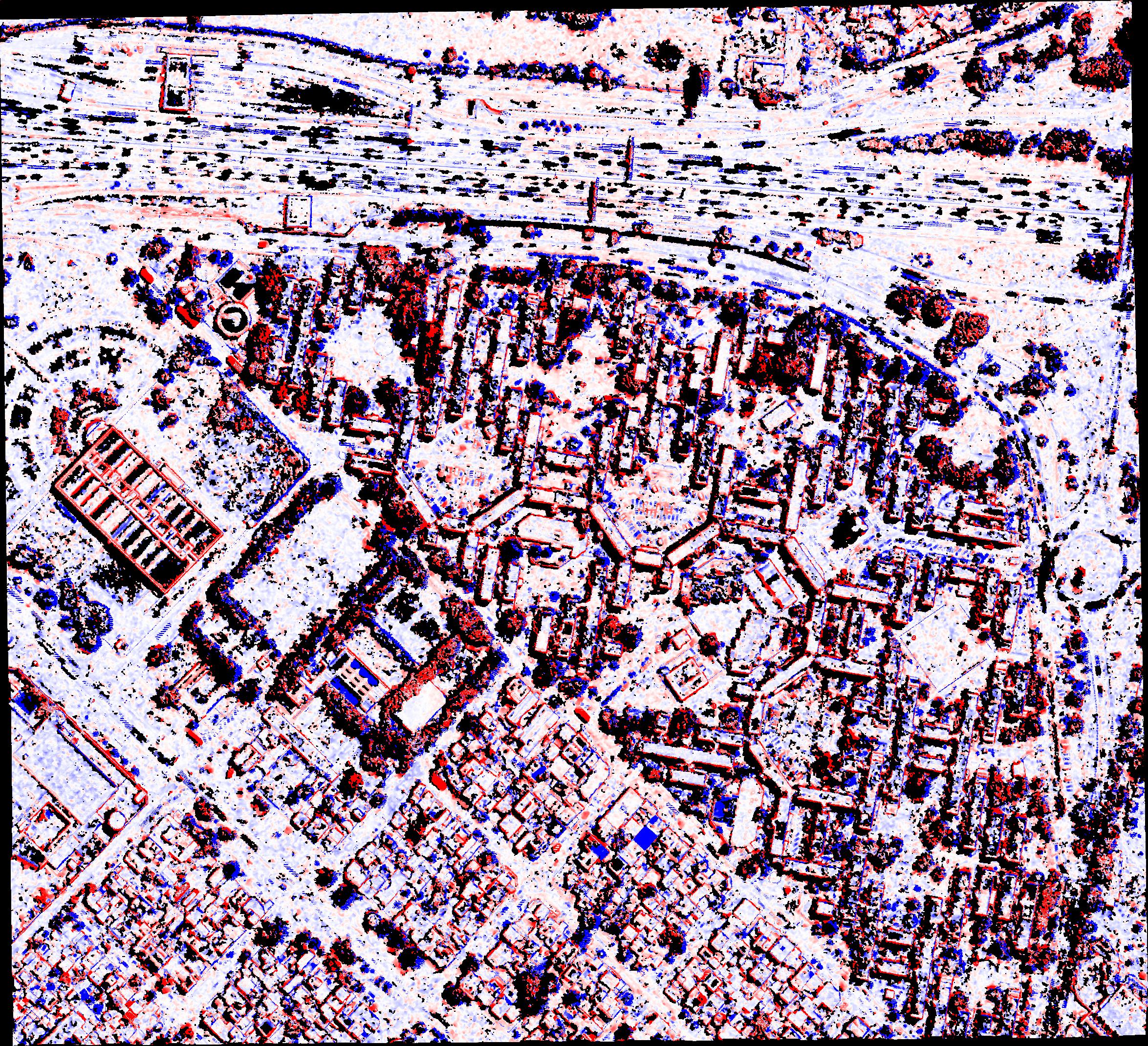} 
    }
    \quad
    \subfigure[S2P17\cite{2017S2P}]{
        \includegraphics[scale=0.05]{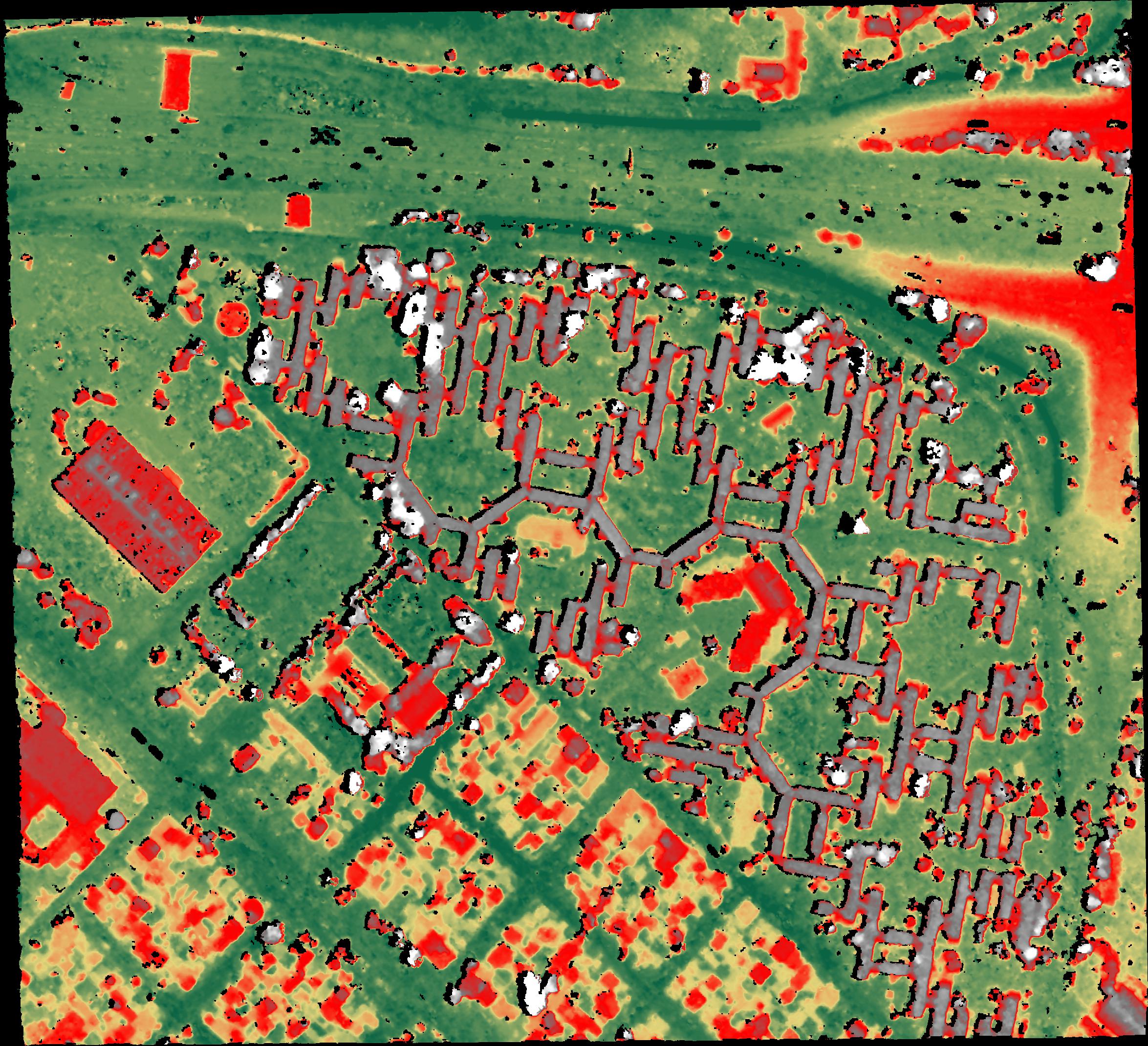} 
        \includegraphics[scale=0.05]{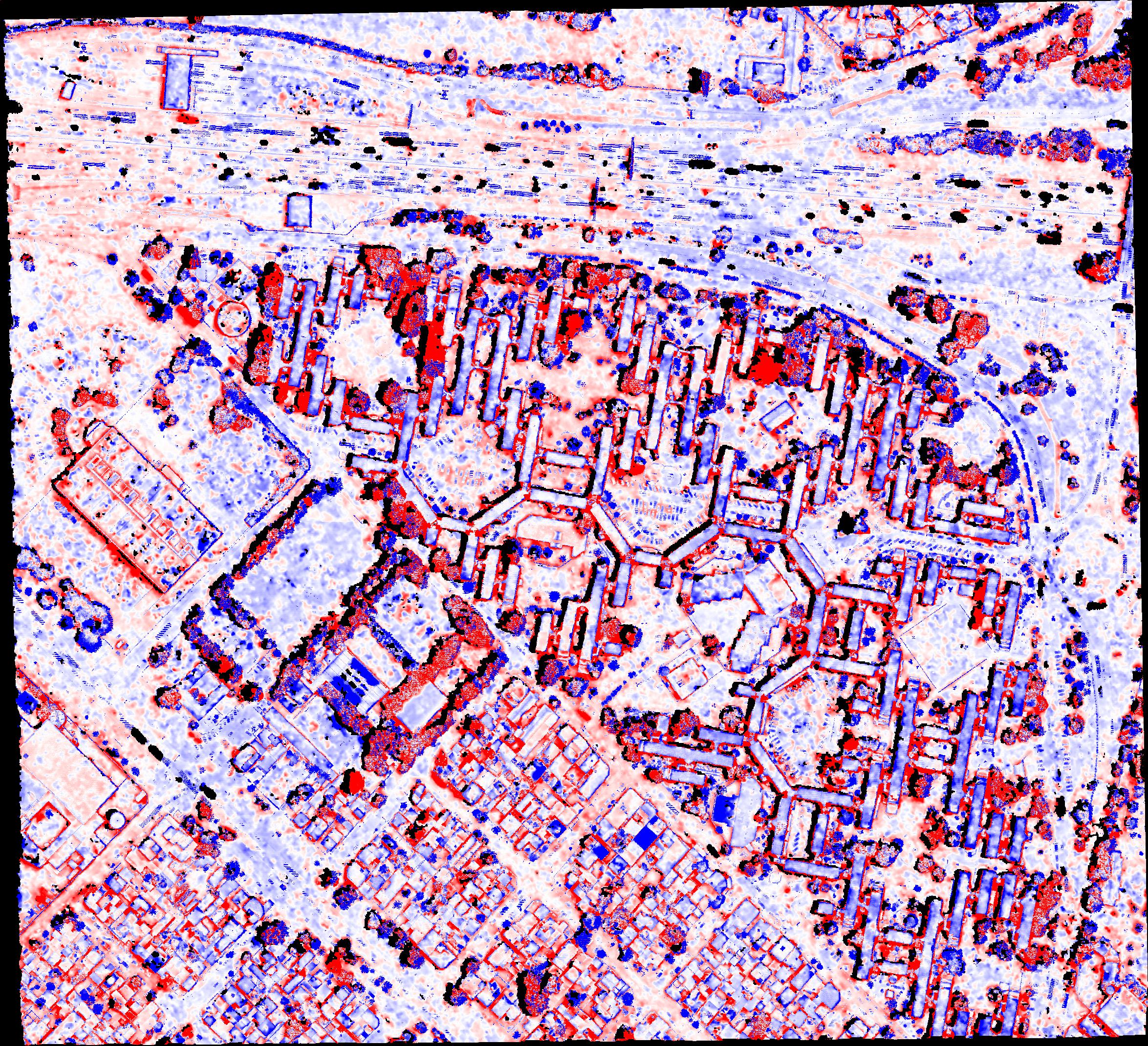} 
    }
    \quad
    \subfigure[Ours]{
        \includegraphics[scale=0.05]{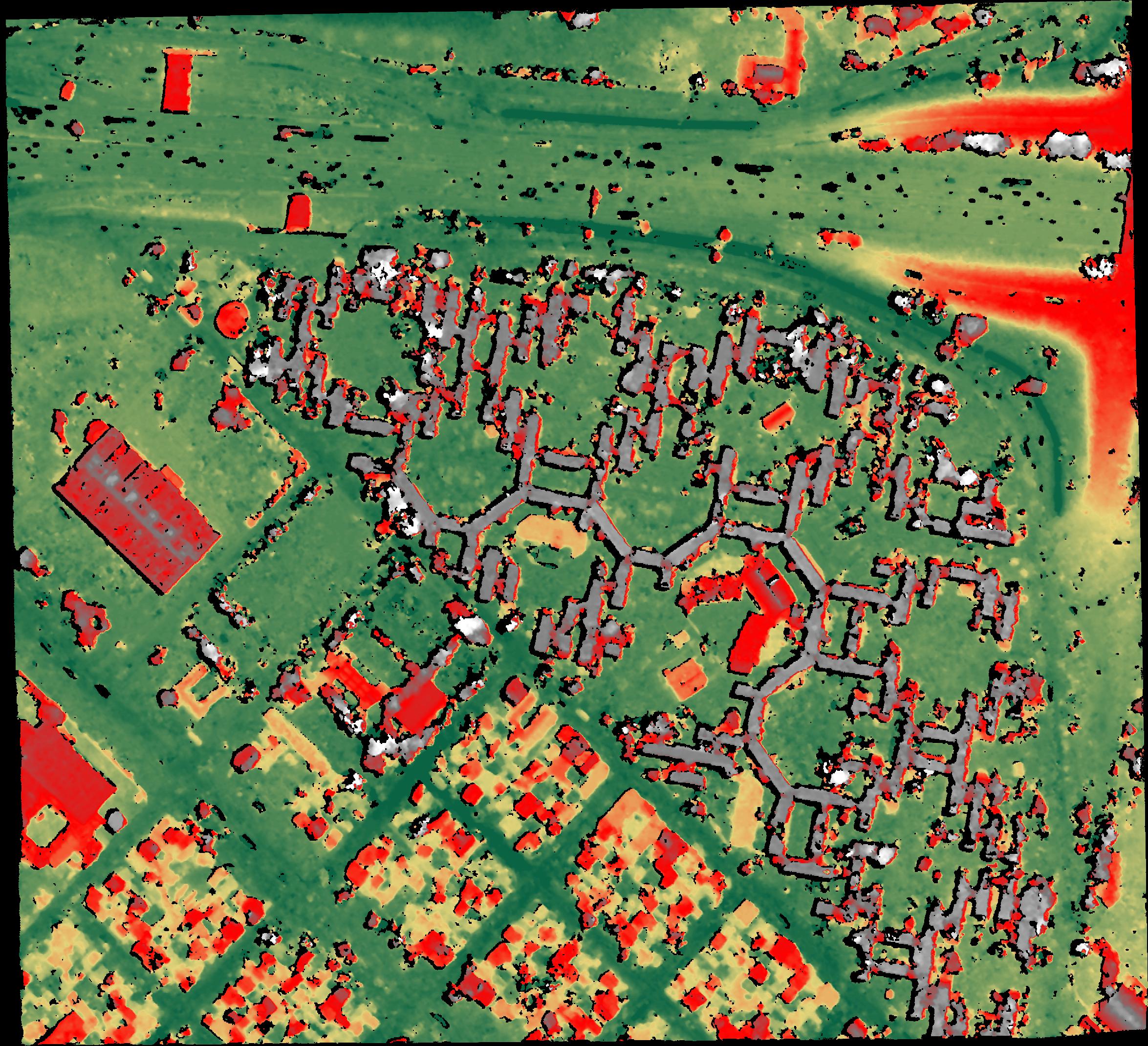} 
        \includegraphics[scale=0.05]{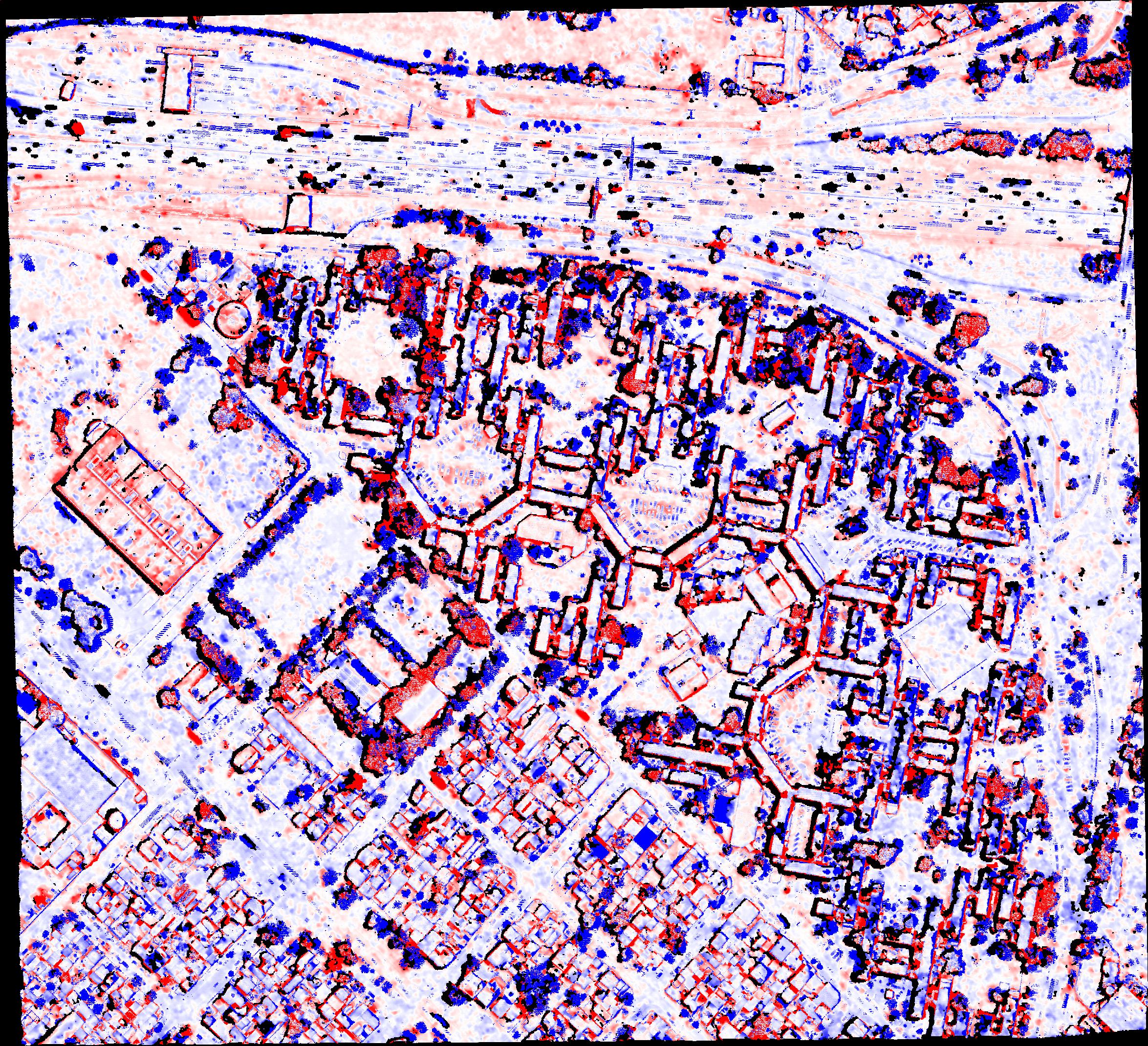} 
    }
    \quad
    \caption{HEIGHT ERROR MAPS ON SITE 1 OF MVS3DM DATASET.}
    \label{fig:my_label}
\end{figure}

\begin{figure}[htbp]
    \centering
    \quad
    \subfigure[Ground-Truth \& COLOR BAR]{
        \includegraphics[scale=0.13]{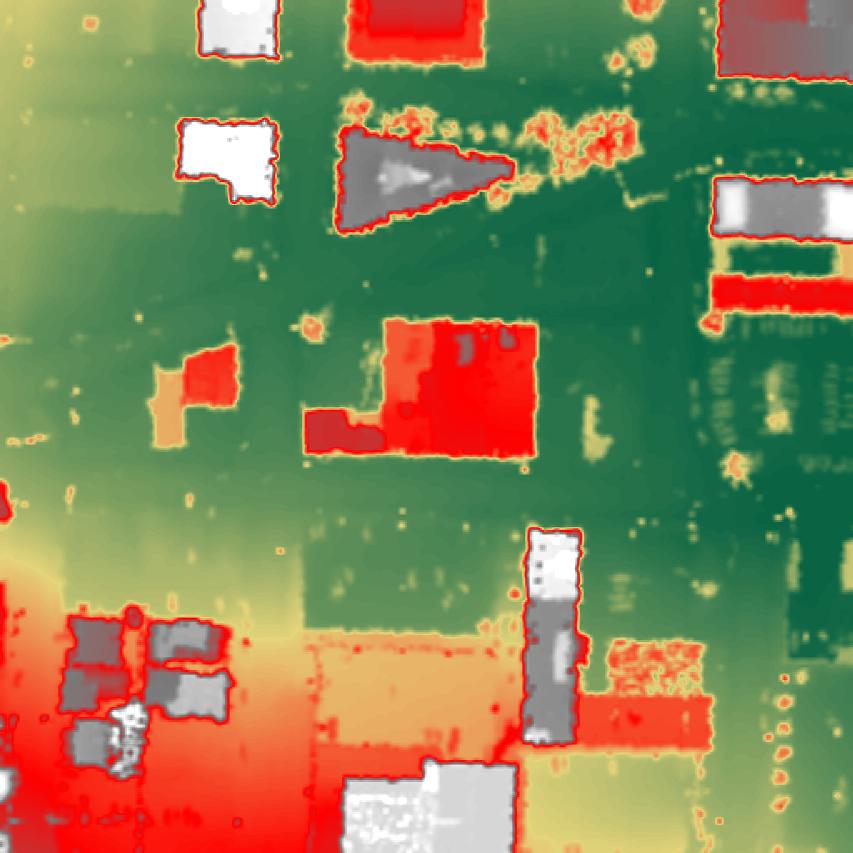} 
        \includegraphics[scale=0.3]{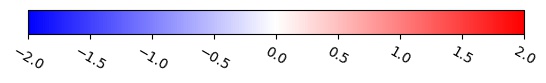} 
    }
    \quad
    \subfigure[JHU/APL\cite{2016BENCHMARK}]{
        \includegraphics[scale=0.13]{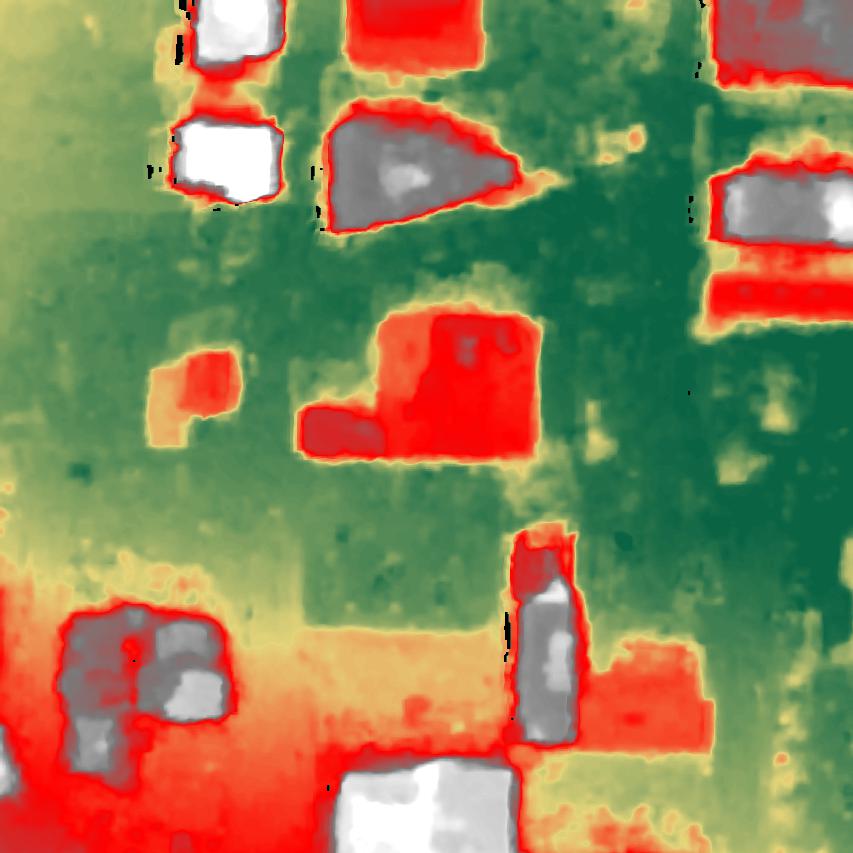} 
        \includegraphics[scale=0.13]{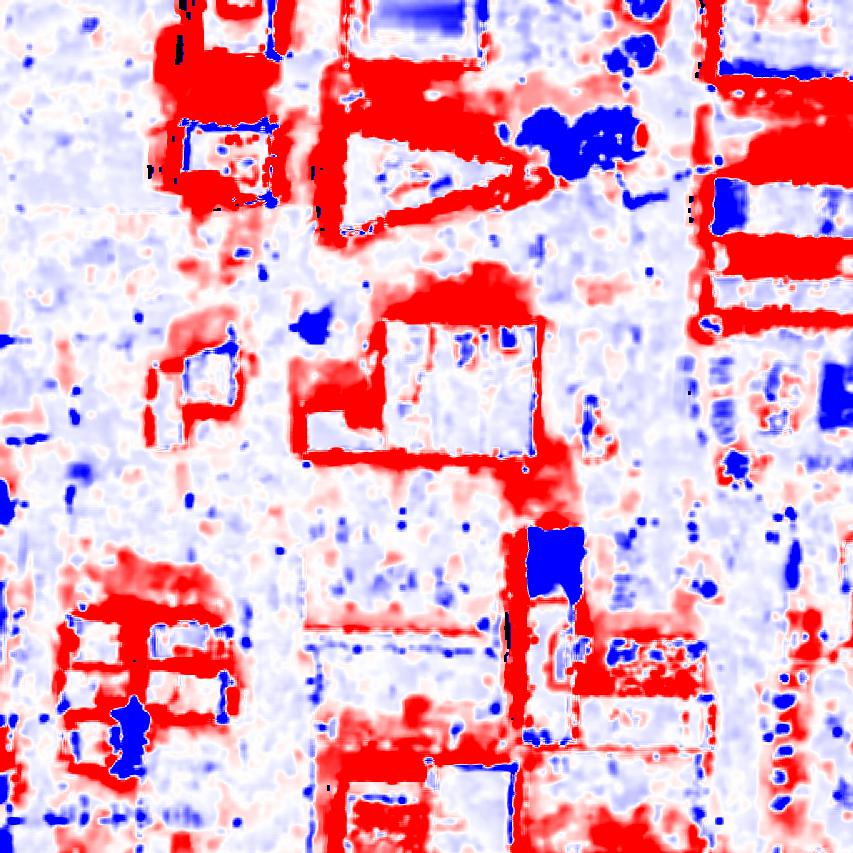} 
    }
    \quad
    \subfigure[COLMAP\cite{2019COLMAP}]{
        \includegraphics[scale=0.13]{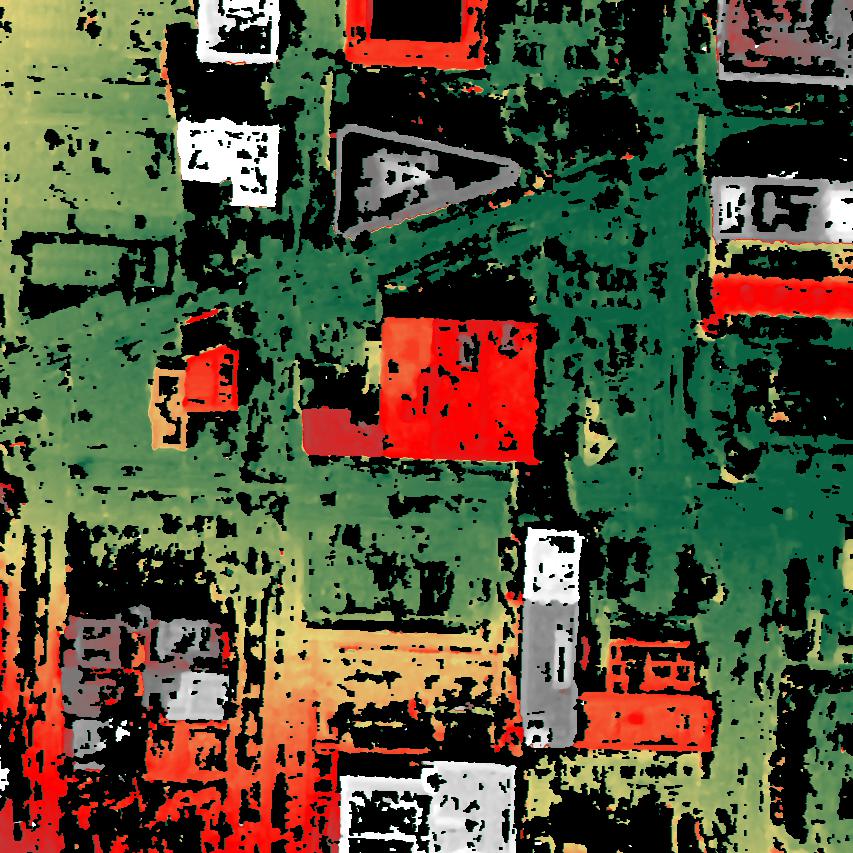}
        \includegraphics[scale=0.13]{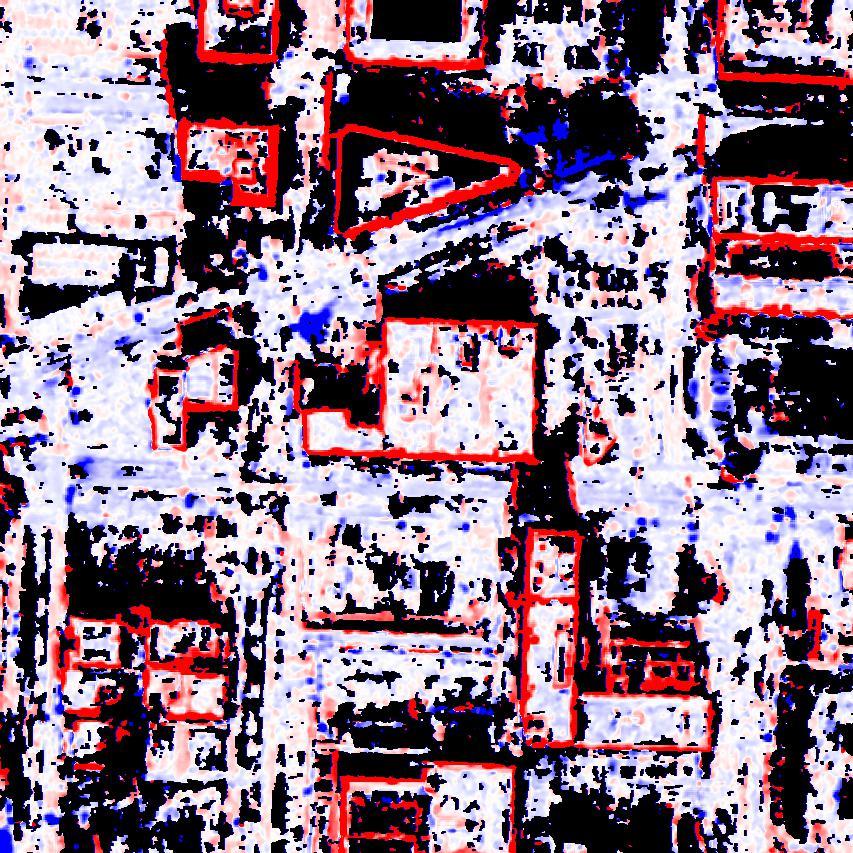} 
    }
    \quad
    \subfigure[S2P17\cite{2017S2P}]{
        \includegraphics[scale=0.13]{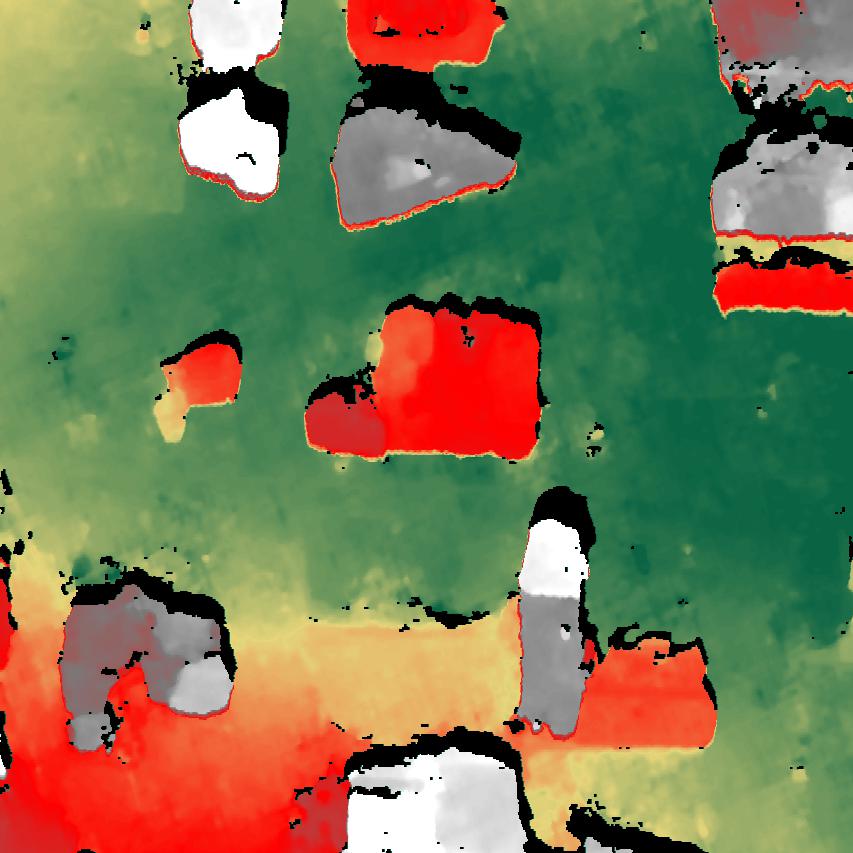} 
        \includegraphics[scale=0.13]{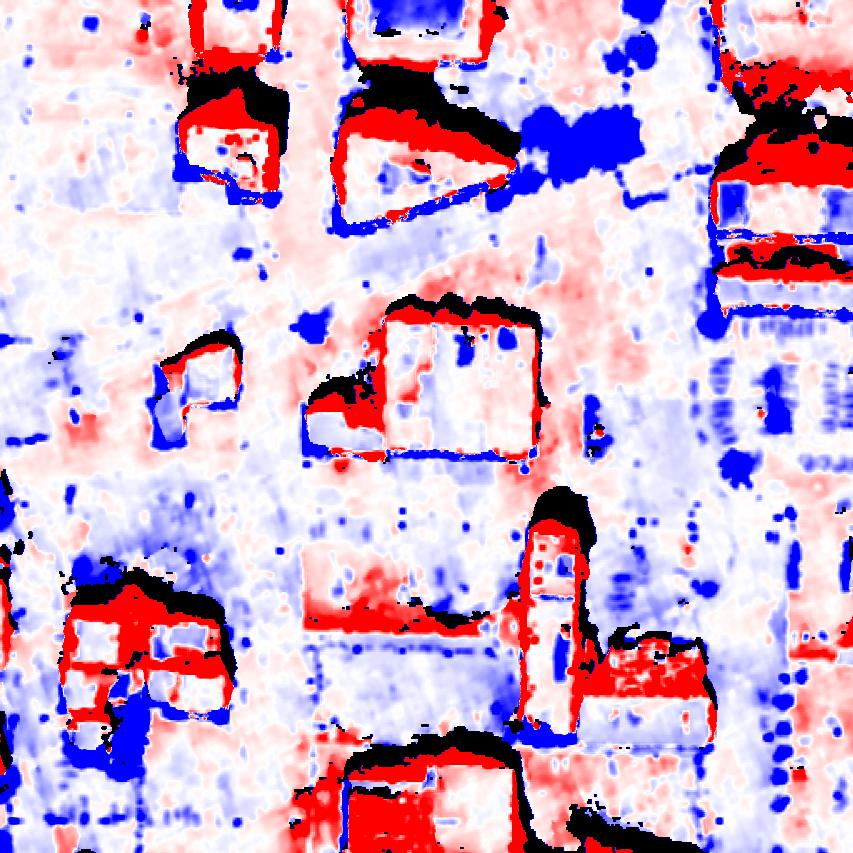} 
    }
    \quad
    \subfigure[Ours]{
        \includegraphics[scale=0.13]{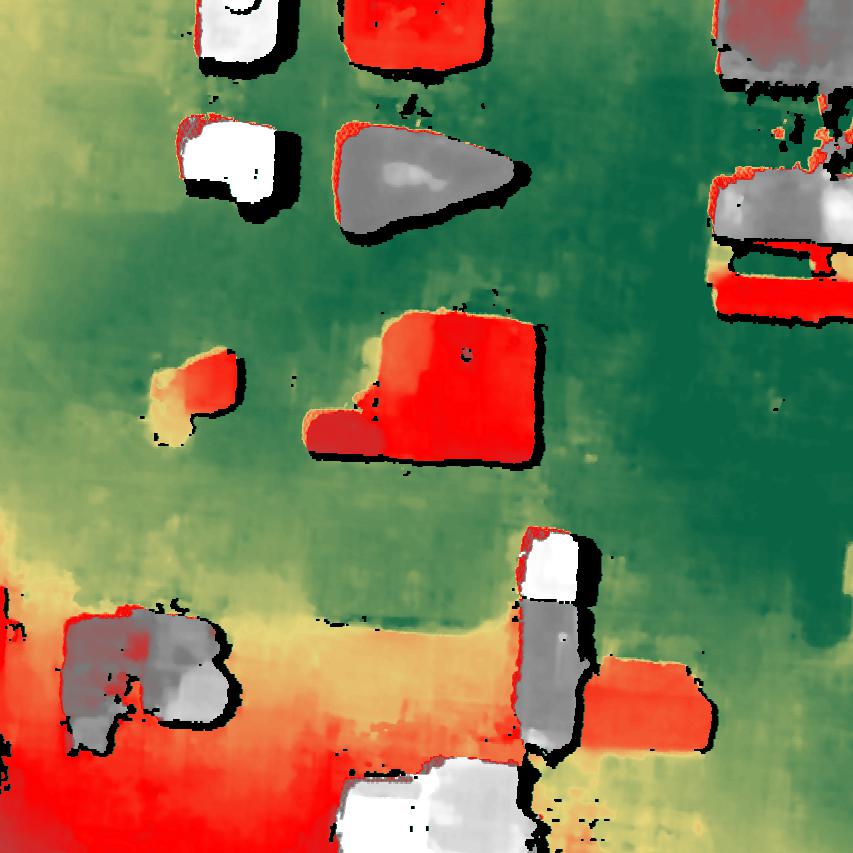} 
        \includegraphics[scale=0.13]{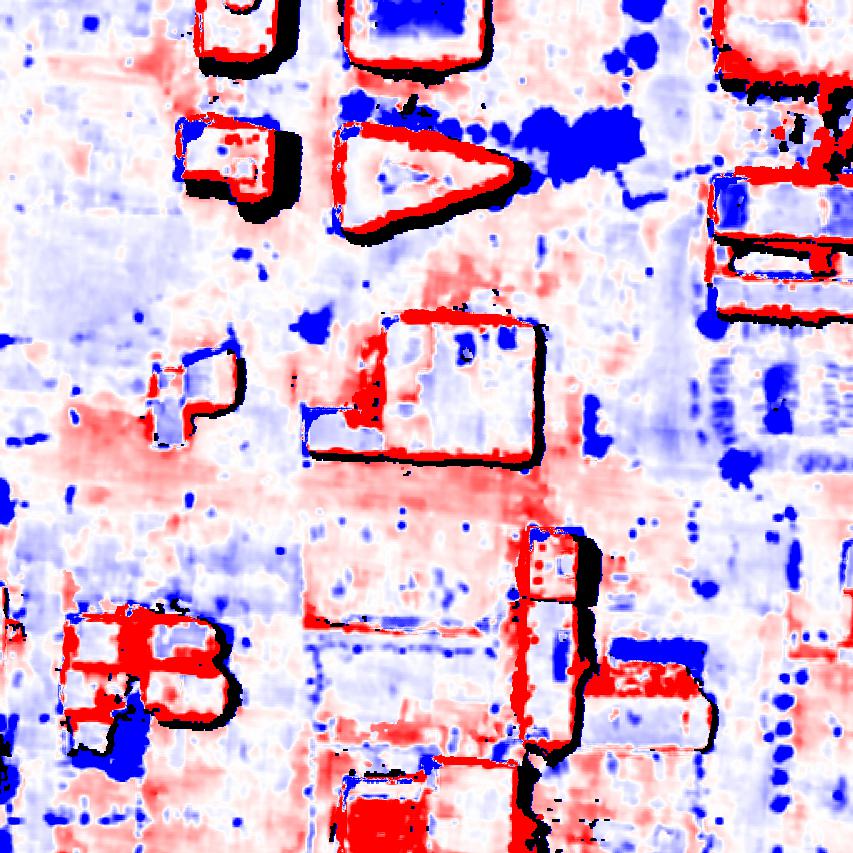} 
    }
    \quad
    \caption{HEIGHT ERROR MAPS ON SITE 6 OF DFC2019 DATASET.}
    \label{fig:my_label}
\end{figure}
\subsubsection{For the complete image set}
In addition, we also report comparative evaluation on the complete image sets corresponding to the first 4 sites in the MVS3DM dataset. Our proposed method performs optional post-processing of image set for each site, and the other three state-of-the-art methods all have their own fusion strategies. As shown in Table 4, our proposed method wins again in the comparison of the reconstruction performance for the entire image set in each site, which is consistent with the evaluation results in Table 2. This is attributed to the fact that our proposed method employs an affine camera model to approximate the small-sized tile in optical satellite image, which not only increase the accuracy of the camera model approximation, but also enables many optimization methods in computer vision filed to be directly applied to our proposed AE-Rec framework. Moreover, the image grouping strategy we designed works well.\\
\indent
However, as can be seen from the last column of Table 4, our proposed method costs more time compared to S2P and COLMAP methods, which is due that we employ an incremental reconstruction strategy in the affine dense reconstruction stage, and thus the iterative local optimization consumes a lot of computation time. S2P \cite{2017S2P} takes the shortest run-time among these methods because it performs only simple linear optimization during the reconstruction. COLMAP \cite{2019COLMAP} run faster than our proposed method in that it only adjusts the position of the principal point in the camera model during bundle adjustment (BA). JHU/APL \cite{2016BENCHMARK} involves solving non-trivial third-order polynomial systems during the RPC model optimization, so it takes the longest run-time. It is necessary to state here that the aforementioned run-time are counted with the RPC model known without considering the run-time required for GCPs collection and RPC model fitting. Otherwise, our method still offers advantages in terms of efficiency. 

\begin{table}[!tb]
\footnotesize 
\caption{COMPARISON OF QUANTITATIVE RESULTS ON COMPLETE SETS OF MVS3DM DATASET.}
\label{tab:Vaihingen_result}
\centering
\begin{tabular*}{\linewidth}{@{\extracolsep{\fill}}@{\quad}l@{\extracolsep{\fill}}@{\quad}c@{\extracolsep{\fill}}@{\quad}c@{\extracolsep{\fill}}@{\quad}c@{\extracolsep{\fill}}@{\quad}c@{\extracolsep{\fill}}@{\quad}c@{\extracolsep{\fill}}@{\quad}c}
\hline
MVS3DM    &   Method  &   CP(\%)    &   ME($m$)  & Time(mins)\\
\hline
\multirow{4}{*}{Site1} & JHU/APL\cite{2016BENCHMARK}    & 72.4 & 0.324 & 151.5\\
                       & COLMAP\cite{2019COLMAP}
  & 72.5 & 0.315 & 86.7\\
                       & S2P\cite{2017METRIC} 
  & 80.1 & 0.317 & \textbf{70.1}\\
                       & Ours 
  & \textbf{81.4} & \textbf{0.215} & 91.3\\	
\hline
\multirow{4}{*}{Site2} & JHU/APL\cite{2016BENCHMARK}    & 70.4 & 0.391 & 147.9\\
                       & COLMAP\cite{2019COLMAP}
  & 66.8 & 0.450 & 83.8\\
                       & S2P\cite{2017METRIC} 
  & 74.0 & 0.486 & \textbf{71.4}\\
                       & Ours 
  & \textbf{75.7} & \textbf{0.345} & 88.2\\	

\hline
\multirow{4}{*}{Site3} & JHU/APL\cite{2016BENCHMARK}    & 59.7 & 0.594 & 127.1\\
                       & COLMAP\cite{2019COLMAP}
  & 63.4 & 0.393 & 50.7\\
                       & S2P\cite{2017METRIC} 
  & 73.1 & 0.524 & \textbf{43.8}\\
                       & Ours 
  & \textbf{74.1} & \textbf{0.379} & 59.5\\	
\hline
\multirow{4}{*}{Site4} & JHU/APL\cite{2016BENCHMARK}    & 43.9 & 1.346 & 119.2\\
                       & COLMAP\cite{2019COLMAP}
  & 50.0 & 1.632 & 48.9\\
                       & S2P\cite{2017METRIC} 
  & 58.6 & 1.091 & \textbf{42.1}\\
                       & Ours 
  & \textbf{60.8} & \textbf{0.856} & 53.9\\	
\hline
\end{tabular*}
\end{table}

\section{Conclusion}
In this paper, for the problem of how to recover the 3D scene structure from optical satellite images using as few GCPs as possible, we propose the AE-Rec method, which consists of two stages: affine reconstruction and affine-to-Euclidean upgrading. Unlike most RPC model-based methods, our method enables to recover the Euclidean structure of multiple optical satellite images with at least 4 GCPs. Experimental results on the MVS3DM and DFC2019 datasets demonstrate the 
advantages of our method compared to the three state-of-the-art optical satellite image 3D reconstruction methods.\\
\indent
In future, we will investigate how to effectively utilize the semantic information in optical satellite images to obtain semantic 3D reconstruction with optical satellite images, and improve the reconstruction accuracy of our proposed AE-Rec framework based on the consistency in the semantic information between images.
\bibliographystyle{ieeetr}
\bibliography{references}
\end{document}